%% file: acl_latex.tex
\documentclass[11pt]{article}

\input{00_decl}

\usepackage[preprint]{acl}

\usepackage{times}
\usepackage{latexsym}

\usepackage[T1]{fontenc}

\usepackage[utf8]{inputenc}

\usepackage{microtype}

\usepackage{inconsolata}

\usepackage{graphicx}

\title{Post-training is (Massive) Supervised Learning}

\author{Michael Hassid$^{1,2}$,  Yossi Adi$^{1,2}$, Roy Schwartz$^{2}$ \\ \\
$^1$FAIR,  Meta Superintelligence Labs \quad
$^2$The Hebrew University of Jerusalem\\ \\
}

\begin{document}
\maketitle
\begin{abstract}
The prevailing paradigm for training LLMs has evolved to rely on a massive post-training phase consisting of SFT and RL. In this position paper, we argue that this methodology effectively marks a reversion to the ``pre-train then fine-tune'' approach of the BERT era, explicitly tailoring models to the desired behaviors and specific benchmarks on which they are evaluated. We begin with a historical overview of LLMs, describing the different phases of the LLM evolution. We argue that the current landscape is remarkably similar to the early days of LLMs, where task performance heavily relied on fitting the models to in-distribution datasets. To empirically demonstrate this, we compare pre-trained models to randomly initialized ones, by fine-tuning both variants on modern reasoning datasets and evaluating them on competitive math and code benchmarks. We show that models post-trained from scratch yield highly non-trivial performance. Our findings suggest that current post-training methodologies function primarily as a distribution-fitting mechanism. We finish by positing that developing generally capable models and systems requires moving beyond extensive post-training for predefined behaviors, shifting instead toward training procedures where models ``learn how to learn''.
\end{abstract}

\input{01_intro}
\input{02_story}
\input{03_exps}
\input{04_discussion}
\input{06_conclusion}

\section*{Limitations}
\label{sec:limitations}

Our experimental scope is subject to two primary limitations. First, our experiments are restricted to reasoning data: competitive math and code. We focused on these areas because they represent a primary focus of the current LLM field~\citep{google_gemini_pro,qwen35blog,Nemotron_Cascade_2} and benefit from the availability of large-scale, open-source SFT datasets~\citep{ocr2,omr,ocr}, while acknowledging that their limited reliance on world knowledge makes them favorable for post-training-only models. Second, our analysis focuses exclusively on SFT, whereas modern post-training pipelines typically include a subsequent RL phase.

Fundamentally, RL requires a pre-training stage for the model's capacity to generate correct solutions to be rewarded on. Coupled with the fact that our pre-trained models consistently outperform their non-pretrained counterparts,  we highlight that pre-training remains critical. Rather, our findings highlight the extent to which current post-training optimizes for specific target distributions, enabling even models initialized from scratch to achieve surprising results on competitive benchmarks.

\section*{Acknowledgments}
We thank Ori Yoran and Gallil Maimon for countless discussions, brainstorming sessions, and their valuable perspectives on this work. We also thank Miri Varshavsky Hassid for the great feedback and moral support.

\bibliography{custom}

\clearpage

\appendix

\input{92_exp_setup}
\input{91_appendix_full_math_results}
\input{94_examples}
\input{93_app_cross_domain}

\input{90_appendix_small_model_results}
\input{95_math_perfomance_training}
\section{AI Assistance Disclosure}
\label{app:ai_disclosure}
We disclose the use of LLMs to assist with text polishing and visualization, with the authors taking full responsibility for the final content.

\end{document}

%% file: 00_decl.tex
\usepackage[utf8]{inputenc} 
\usepackage[T1]{fontenc}    
\usepackage[hidelinks]{hyperref}       
\usepackage{url}            
\usepackage{booktabs}       
\usepackage{amsfonts}       
\usepackage{nicefrac}       
\usepackage{microtype}      
\usepackage{xcolor}         

\usepackage{microtype}
\usepackage{booktabs}

\usepackage{amssymb}
\usepackage{amsmath}
\usepackage{tabularx}
\usepackage{multirow}
\usepackage{graphicx}
\usepackage{subcaption}
\usepackage{caption}

\usepackage{algorithm}
\usepackage{booktabs}       
\usepackage{diagbox}
\usepackage{xspace}
\usepackage{listings}
\usepackage{wrapfig}
\usepackage{makecell}
\usepackage{calc}

\usepackage[normalem]{ulem}

\usepackage{nicefrac}
\graphicspath{ {./graphs/} }
\usepackage{pgf}

\usepackage[capitalize,noabbrev]{cleveref}

\newcommand{\com}[1]{}
\newcommand{\resolved}[1]{}

\usepackage{etoolbox}
\newtoggle{release}

\toggletrue{release}

\iftoggle{release}{
\newcommand{\michael}[1]{}
\newcommand{\roy}[1]{}
\newcommand{\adios}[1]{}
}
{
\newcommand{\michael}[1]{{\textcolor{orange}{\textbf{Michael:} {#1}}}}
\newcommand{\roy}[1]{{\textcolor{blue}{\textbf{Roy:} {#1}}}}
\newcommand{\adios}[1]{{\textcolor{magenta}{\textbf{YA:} {#1}}}}
}

\newcommand{\ros}[1]{\textcolor{red}{\sout{#1}}}

\renewcommand{\ros}[1]{}

\usepackage[many]{tcolorbox}
\usepackage{xcolor}

\definecolor{systemgray}{HTML}{F4F4F4}
\definecolor{userblue}{HTML}{E8F0FE}
\definecolor{modelpurple}{HTML}{F3E5F5}
\definecolor{bordergray}{HTML}{B0B0B0}

\newtcolorbox{promptbox}[2][]{%
  enhanced,
  boxrule=0.5pt,             
  colframe=bordergray,       
  colback=#2,                
  arc=3pt,                   
  boxsep=2pt,
  left=6pt, right=6pt, top=4pt, bottom=4pt,
  fonttitle=\small\bfseries\sffamily,
  coltitle=black,
  attach boxed title to top left={yshift=-1mm, xshift=3mm},
  boxed title style={
    colback=white,
    boxrule=0.5pt,
    colframe=bordergray,
    arc=2pt,
    top=1pt, bottom=1pt, left=3pt, right=3pt
  },
  title=#1,                  
  #1                         
}

%% file: 01_intro.tex
\section{Introduction}
\label{sec:introduction}

Historically, the prevailing machine learning paradigm relied on a fully supervised approach~-~training models directly on data distributions that corresponded to target tasks. This methodology was overtaken by the self-supervised approach that became dominant with large language models (LLMs). Over the years, the LLMs training paradigm has evolved to incorporate a post-training stage, consisting of supervised fine-tuning (SFT) and reinforcement learning (RL). These steps were designed to align the model towards specific behaviors. Today, this phase is a massive part of the training pipeline, consuming up to hundreds of billions of tokens \citep{olmo_3, Nemotron_Cascade_2,qwen_coder_next}.

\begin{figure}[t]
    \centering
    \includegraphics[trim={0.1cm 6.64cm 2.25cm 0.0cm},clip,width=1\linewidth]{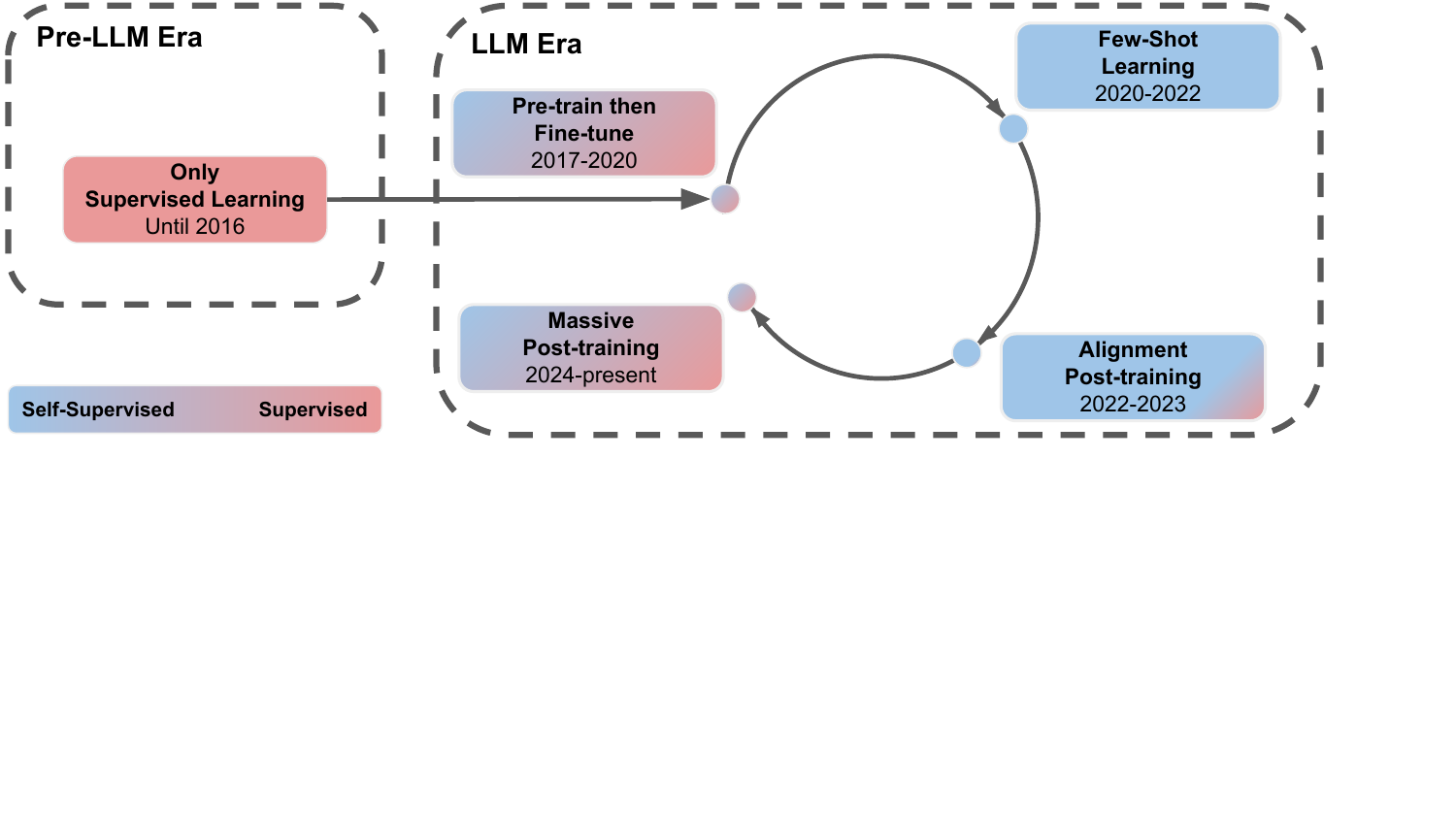}
    
    \caption{
    Evolution of LLM training paradigms. The field has progressed from only supervised learning (pre-LLM era), through several LLM-based stages: ``pre-train then fine-tune'', few-shot learning, and post-training for alignment. 
    We argue that the current era of massive post-training brings the field (almost) full circle. 
    By heavily shifting model distributions toward specific benchmarks and 
    desired behaviors, modern post-training effectively reverts to the supervised, distribution-fitting, mechanisms of earlier LLM eras.
    }
    \label{fig:fig_1}
\end{figure}

In this position paper, we argue that the NLP field has come full circle (\cref{fig:fig_1}). Today, the training of leading LLMs, where post-training has become such a critical component, has effectively transformed back into a form of supervised learning. This paradigm marks a reversion to the ``pre-train then fine-tune'' methodology characteristic of the BERT era~\citep{bert, elmo, gpt1}. Rather than pursuing true zero-shot generalization \citep{gpt3}, the prevailing strategy has shifted back toward multi-task optimization, explicitly fitting models to the distributions on which they are evaluated.

We begin with a brief historical overview of LLMs. We start by charting their early days as small-scale backbones for supervised learning models \citep{bert,roberta}. We then trace the shift toward massive pre-training scale, where models began achieving high performance via in-context learning with only a handful of inference-time examples~(few-shot learning; \citealp{gpt3}). Next, we describe the evolution of the post-training phase: what began as a short stage aimed at aligning a model's behavior and formatting has expanded into a massive phase using hundreds of billions of tokens for SFT and RL trainings. We follow with a short discussion of the differences between the two approaches. We conclude by arguing that, by systematically shifting model distributions toward specific benchmarks and desired behaviors, current methodologies effectively transform evaluation distributions into explicit training distributions.

To empirically demonstrate this observation, we design a series of experiments isolating the impact of large-scale post-training (specifically SFT) from traditional pre-training. We posit that if post-training merely unlocks or aligns pre-existing knowledge acquired during pre-training, a model trained from scratch strictly on SFT data should fail on reasoning tasks. Conversely, if it acts as a distribution-fitting mechanism, the non-pretrained model should achieve non-trivial performance.

We evaluate models across varying parameter scales on mathematical reasoning and competitive programming benchmarks. Remarkably, our results validate the latter scenario, demonstrating that SFT enables models to achieve surprisingly high results even in the complete absence of traditional pre-training. We find that models trained entirely from scratch, using SFT datasets only and without prior linguistic or domain exposure, manage to achieve non-negligible results on highly difficult benchmarks. E.g., a 14B model scores 51\% on LCB V5 \citep{lcb} and a 7B model obtains 43\% on AIME-24 \citep{matharena}. Importantly, these models underperform their pre-trained counterparts by a margin of 8–20\%, gaps that mirror the BERT and GPT-1 era, where pre-trained models held a similar advantage over supervised-only baselines.
Furthermore, we show that a minimal pre-training setup of less than 0.1\% of the original pre-training compute bridges up to 84\% of the remaining performance deficit between non-pretrained and fully pre-trained models.
Ultimately, this suggests modern post-training depends far less on pre-training's emergent capabilities than previously thought, as high performance on desired benchmarks can rely on pulling targeted tasks directly into the model's distribution.

We conclude that current LLM development relies on a massive, yet fundamentally reactive, post-training paradigm. Because this approach is inherently reactive, the field is trapped chasing the ever-evolving demands of LLM applications.
We suggest that the path to generally capable models requires moving beyond static post-training for predefined benchmarks. We advocate for a shift in the training procedures, and suggest that they should include meta-learning where models \textit{learn how to learn}. 
This will lead toward models and agentic systems designed to inherently self-evolve to solve novel, out-of-distribution tasks~\citep{autoflow,evolveR,recursive_lms,survey_1,survey_2}. 

As a position paper, our contributions are not a proposal of a new method or training paradigm. Instead, our goal is to provide a critical perspective on the current state of LLM development, particularly the growing role of post-training in shaping model capabilities. Through both historical analysis and empirical observations, we argue that modern practices increasingly resemble traditional supervised learning at scale and distribution-fitting approaches. We highlight the broader implications of this trend for the field, and propose a high level direction for future research.

%% file: 02_story.tex
\section{Back to Supervised Learning}
\label{sec:story}

We start by reviewing the history of LLMs, from their early days as backbones for fine-tuning to  the modern, massive, post-training pipelines that characterize current models.~(\cref{subsec:history_of_llms}). Next, we discuss the differences between modern post-training methods and traditional ones~(\cref{subsec:trad_vs_modern}).

\subsection{Pre-training and Post-training in LLMs: A Historical Overview}
\label{subsec:history_of_llms}

\paragraph{From supervised learning to pre-train then fine-tune.} 
The prevailing approach prior to the current era of LLMs involved a straightforward supervised learning pipeline \citep[\textit{inter alia}]{eisner1996three,klein-manning-2003-accurate,shalev2014understanding,fast_parsing,wang-etal-2016-attention,seo2016bidirectional}. To tackle a specific task, researchers would first gather data and create a labeled dataset, then partition it into training and testing subsets. A model was subsequently trained on the training data, and its performance was evaluated on the validation and test sets. These evaluation sets were inherently highly in-distribution, meaning they were highly similar to the training data. 

The introduction of models like ELMO, BERT, and GPT-1 marked a paradigm shift with the advent of the modern \textit{pre-training} phase~\citep{elmo,bert,gpt1}. This new methodology involved first training the model in a self-supervised manner on large volumes of general text data, using techniques such as (masked) language modeling. These pre-trained models were subsequently fine-tuned on task-specific training sets and evaluated on their corresponding test sets~\citep{glue,superglue}. This fine-tuning approach yielded performance improvements of up to 20\% over their supervised-only counterparts~\citep{elmo,gpt1}.
This ``pre-train then fine-tune'' approach quickly became the new standard for training LLMs~\citep{roberta, electra, t5}, and was also adapted for other domains like vision and audio~\citep{audio_survey,audio_survey_2,vision_survey}.

\paragraph{Scaling and few-shot learning.}
Subsequent research focused on progressively increasing the scale of this pre-training phase, eventually making it the sole dominant component of the training process. A notable example is GPT-3 \citep{gpt3}, which was exclusively pre-trained and demonstrated advanced few-shot capabilities. This showed that models could achieve strong performance on benchmarks even without being explicitly trained on those specific data distributions. This approach became the de-facto approach for the following couple of years~\citep{jurassic,opt, bloom}.

\paragraph{Post-training as a tool for model alignment.} Although large-scale pre-trained models demonstrated impressive capabilities on leading benchmarks, they often failed to follow explicit instructions or exhibit human-desired behaviors. 
To bridge this alignment gap, a post-training stage based on SFT and RL was established. Using a relatively small amount of curated data, this phase aligned a general-purpose base model to instruction and chat formats \citep{flan,rlhf}, as well as to human preferences \citep{rlaif}. This preference alignment is typically achieved through RL frameworks (such as RLHF; \citealp{rlhf_1,rlhf_2}). 

\paragraph{New, hard benchmarks arise.}
With the advancement of LLM performance, the research community started building harder and more complicated benchmarks, with the goal of pushing the boundaries of the LLM technology.
Early benchmarks such as GSM8k for mathematics and HumanEval for coding \citep{gsm8k,humaneval}, targeted tasks that were challenging for models but relatively simple for humans. However, the ceiling for evaluation has since been raised by math competition benchmarks \citep{matharena}, like AIME and HMMT, live code bench (LCB; \citealp{lcb}), which tests competitive programming, and SWE/Terminal-bench which tests comprehensive software engineering skills \citep{swebench,swe_pro,ma2025swe,yang2025codeclash,press2025algotune,terminal_bench}. These newer benchmarks test LLMs on tasks requiring human expert capabilities.

\paragraph{The LLM community response: Post-training.}
As the difficulty of benchmarks increased, the LLM community responded by tailoring models to these specific performance metrics~\citep{claude_35, gpt_5,kimi_25,Nemotron_Cascade_2,research2026composer}.  This has led to a post-training phase where the model's data distribution is intentionally shifted to align with the desired output distribution, shaped by both the benchmark requirements and human preference data \citep{terminal_data,open_swe}. 

Consequently, post-training has evolved into a pivotal stage of the development lifecycle. This phase typically employs SFT datasets to bootstrap the RL process, targeting high-value tasks~—~such as software engineering \citep{swebench, swe_pro}, command-line control \citep{terminal_bench}, multi-hop web reasoning \citep{hle}, and more \citep{gpqa, tau2, arc}. Scaling these post-training processes has emerged as a dominant paradigm. SFT data volumes have expanded up to hundreds of billions of tokens, incorporating diverse sources such as OMR, OCR, Nemotron-Math, Terminal-Corpus and more \citep{omr, ocr, ocr2, nemotron_math, endless, terminal_data}. This trajectory extends to the subsequent RL phase, which now scales to thousands of RL environments and hundreds of billions of interaction tokens~\citep{swe_smith, olmo_3,swe_rebench,Nemotron_Cascade_2,open_swe}. Notably, current post-training pipelines now process data at a scale equivalent to the entire pre-training phase of models from just four to five years ago, such as GPT-3 and OPT~\citep{gpt3, opt}. \looseness=-1

\paragraph{Back to in-distribution specialization.} We argue that the growth of post-training suggests a skepticism regarding the few-shot generalization capabilities of pre-training alone~\citep{gpt3}. By systematically shifting model distributions toward specific benchmarks and desired behaviors, current methodologies effectively transform the evaluation distributions into explicit training distribution \citep{soft_cont}.

In that sense, the NLP field has come full circle (\cref{fig:fig_1}). Current LLMs are reminiscent of early LLMs like GPT-1 and BERT: we have pre-trained models that we fine-tune using supervised learning over specific distributions, which are the relevant distributions on which we evaluate them. 
This trajectory effectively marks a reversion to the ``pre-train then fine-tune'' methodology. Analogous to the BERT era, the prevailing strategy has shifted away from zero-shot generalization and back toward specific multi-task optimization, where models are explicitly fitted to the specific distributions on which they are evaluated and \textit{currently} used. 
While these strategies may generalize to distributions similar to the training set, they often fail to generalize to underrepresented or significantly divergent distributions~\citep{esolang}.

This methodology has established a recurring cycle. When a new capability is required, a benchmark is introduced, typically one where initial model performance is much worse compared to humans. Yet, these benchmarks are often saturated within a short time (a year or so). This rapid progress is frequently driven by the post-training process, where the benchmark is effectively pulled in-distribution via the curation of analogous SFT datasets and the construction of proxy RL environments. \looseness=-1

\subsection{Traditional vs.~Modern Post-training}
\label{subsec:trad_vs_modern}

While the current LLM development pipeline marks a philosophical reversion to the ``pre-train then fine-tune'' methodology characteristic of the BERT era, critical distinctions between the two paradigms remain. Below, we discuss several key aspects of these differences.

Historically, the traditional post-training (or fine-tuning) phase was a relatively short stage. It used minimal data compared to the pre-training phase, serving merely to adapt a backbone model to a specific format. During the BERT era, it was mainly used for training on the specific training examples of each dataset \citep{bert,roberta}, while in the subsequent instruction-tuning era, it used millions of tokens~\citep{rlhf,llama_2}. 
In contrast, modern post-training has expanded into a massive phase. Current pipelines process diverse data sources at a scale of hundreds of billions of tokens for SFT and extensive RL \citep{qwq,kimi_25,cwm,Nemotron_Cascade_2}.
Notably, this phase now processes data volumes equivalent to the entire pre-training runs of leading models from just four to five years ago, such as GPT-3 and OPT~\citep{gpt3,opt}.

Furthermore, these traditional and modern paradigms differ in their data-generation mechanisms. Traditional post-training relies exclusively on static, task-specific datasets~\citep{glue,superglue}. In contrast, modern post-training employs a more dynamic approach. The SFT phase remains analogous to traditional supervised learning, optimizing over explicit input-output pairs (often distilled from other models). However, the RL phase shifts the supervisory signal exclusively to the terminal output (e.g., via a reward model); consequently, the intermediate training trajectories are generated dynamically by the model actively undergoing training. 

A related mechanical distinction lies in the formulation of the training objective. Traditional fine-tuning framed downstream evaluations as classification or extraction tasks, typically applying the loss over a single output token—such as a dedicated classification head~\citep{glue,superglue}. Modern post-training, by contrast, is fundamentally generative. During SFT, the model optimizes next-token prediction over the output trajectory; while during RL, it optimizes a reward-based objective over generated trajectories.

Importantly, similarly to the traditional fine-tuning setup, the whole modern post-training paradigm can be categorized as a supervised learning phase. Given that both SFT and RL ultimately rely on annotated data, the former through direct label mapping and the latter through the evaluation of model outputs, the post-training phase is guided by task-specific supervision.

Another difference lies in the relationship between training and test datasets. In the classic supervised pipeline, where each task was evaluated individually, a dataset was first collected and then randomly partitioned into training and test subsets.
Because they originated from the exact same source, the test set was inherently highly in-distribution. Today, this dynamic is entirely inverted. The modern cycle begins with the introduction of a novel, highly difficult benchmark (the test set), where initial model performance is poor. To solve it, the LLM community systematically mines, synthesizes, and curates analogous SFT datasets and proxy RL environments designed to closely resemble the evaluation criteria~\citep{omr,ocr,Nemotron_Cascade_2}. Despite these differences, the result is similar: in both setups the training distribution closely resembles the test distribution.

Finally, a prominent distinction between modern post-training and the traditional approach is that the latter typically focused on a single task. In contrast, modern post-training objectives aim for the development of unified, generalist models by leveraging \emph{multi-task} environments. This simultaneous exposure to various data distributions is intended to promote broader generalization beyond the explicit post-training data. For example, a model post-trained on competitive programming in a primary language is expected to generalize to alternative languages, assuming those secondary languages maintain some representation within the broader corpus. Rather than producing a collection of models for each task, this multi-task training yields a single, \emph{general-purpose} model capable of synthesizing these learned capabilities to excel across diverse tasks.

In the following section, we show that modern post-training (specifically, large-scale SFT) can yield non-negligible results completely independent of pre-training. This highlights just how heavily current post-training pipelines shift model distributions to fit specific tasks.

%% file: 03_exps.tex
\section{Massive SFT can do without Pre-training}
\label{sec:exps}

In this section, we investigate the efficacy of large-scale post-training in isolation. We empirically demonstrate that training LLMs on current SFT datasets yields surprisingly high in-distribution performance on hard math and code benchmarks, even in the absence of pre-training. This finding suggests that current post-training methods heavily optimize models to excel on specific benchmarks, and that certain desired behaviors may rely less on the pre-training phase than previously thought.

We outline our experimental setup in \cref{subsec:exp_steup}. Next, we detail our findings in \cref{subsec:results}. We discuss our experimental scope and limitations in \cref{sec:limitations}. \looseness=-1

\subsection{Experimental Setup}
\label{subsec:exp_steup}

To understand the models' performance of pre-trained vs.~scratch initialization using SFT on desired distributions, we focus on competitive code and math benchmarks. We train models with a range of sizes and compare the results between the pre-trained versions to their from-scratch counterparts. 

We train both variants on two distinct, large-scale SFT reasoning datasets. The math models are trained on the Open Math Reasoning dataset (OMR; \citealp{omr}), consisting of $\sim$3.2M question-answer pairs. 
The code models are trained on the Open Code Reasoning II dataset (OCR-II; \citealp{ocr2}), which contains about 1.4M Python and 1.2M C++ question-answer pairs.

We compare the different models' performance on  highly tracked math and competitive programming benchmarks to understand their reasoning capabilities~—~which is a primary focus and desired proficiency  of recent model releases \citep{google_gemini_pro,qwen35blog,Nemotron_Cascade_2}. Our math evaluation uses MATH-500~\citep{math500}, and four competition math benchmarks from the MathArena leaderboard which are used as the current standard for testing leading math reasoning models~\citep{matharena}: AIME 24, AIME 25, HMMT February 25, and HMMT November 25. To evaluate reasoning for coding, we test the models over the LCB coding benchmark \citep{lcb}, using the V5 and V6 splits. We provide further details about the benchmarks in~\cref{app:exp_setup}.

We emphasize that both OMR and OCR-II were decontaminated against the evaluation benchmarks during their construction. OMR uses LLM-based comparisons to remove questions that closely resemble math benchmark problems~\citep{omr}, while OCR-II uses similarity search followed by LLM-based judging to filter potentially overlapping code questions, including overlap with LCB~\citep{ocr2}.

To isolate the impact of pre-training on specific target distributions~-~in our case, reasoning math and code~-~we focus on models that lack explicit reasoning data (i.e., thinking trajectories) in their pre-training stage. Specifically, we use the Qwen-2.5-Base family of models, which offers extensive training on 18 trillion tokens, strong performance, and availability across multiple scales \citep{qwen_25}. Our experiments use the 1.5B, 3B, 7B, and 14B parameter sizes.

We train all models on each of the SFT datasets for 50k steps, using a sequence length of 32k and a batch size of $\sim$2M tokens~(a total of $\sim$100B tokens). Models trained from scratch use depth-scaled initialization~\citep{depth_init} and retain the original tokenizer, which can theoretically carry priors from its construction. For evaluation, we sample 10 answers per question~(temperature = 0.6, top-p = 0.95) and report both pass@1~(average accuracy) and pass@5 (success rate within five attempts). See~\cref{app:exp_setup} for more details.

\subsection{The Efficacy of SFT-only over Non-Pre-trained Models}
\label{subsec:results}

\begin{table}[t!]
\centering
\tiny
\setlength{\tabcolsep}{4.4pt}
\begin{tabular}{l|l||cc||cc}
\toprule & & \multicolumn{2}{c||}{\multirow{2}{*}{\small\textbf{Math Models}}} & \multicolumn{2}{c}{\multirow{2}{*}{\small\textbf{Code Models}}} \\ & & \multicolumn{2}{c||}{} & \multicolumn{2}{c}{} \\ \textbf{Size} & \textbf{Model Type} & \textbf{MATH-500} & \textbf{Comp.\ Math} & \textbf{LCB V5} & \textbf{LCB V6} \\ & & {pass@1\,|\,5} & {pass@1\,|\,5} & {pass@1\,|\,5} & {pass@1\,|\,5} \\ \midrule
\multirow{5}{*}{1.5B} & Base (no SFT) & \phantom{0}2.4\,|\,10.0 & \phantom{0}0.0\,|\,\phantom{0}0.0 & \phantom{0}0.8\,|\,\phantom{0}3.0 & \phantom{0}0.7\,|\,\phantom{0}3.0 \\
 & Base (Relaxed) & \phantom{0}8.7\,|\,28.5 & \phantom{0}0.1\,|\,\phantom{0}0.4 & \phantom{0}1.6\,|\,\phantom{0}5.4 & \phantom{0}1.7\,|\,\phantom{0}5.7 \\
 & 0\%-Pre & 74.4\,|\,87.6 & 24.7\,|\,43.4 & 35.0\,|\,52.1 & 33.6\,|\,48.8 \\
 & 0.1\%-Pre & 70.4\,|\,84.0 & 18.8\,|\,34.2  & 29.7\,|\,46.6 & 30.2\,|\,45.1 \\
 & 100\%-Pre & 90.1\,|\,96.8 & 38.8\,|\,61.4 & 49.2\,|\,64.7 & 47.4\,|\,61.2 \\
\midrule
\multirow{5}{*}{3B} & Base (no SFT) & \phantom{0}4.1\,|\,14.8 & \phantom{0}0.5\,|\,\phantom{0}2.1 & \phantom{0}4.0\,|\,10.3 & \phantom{0}5.2\,|\,12.3 \\
 & Base (Relaxed) & 16.1\,|\,44.5 & \phantom{0}0.7\,|\,\phantom{0}2.8 & \phantom{0}4.5\,|\,10.4 & \phantom{0}5.4\,|\,12.3 \\
 & 0\%-Pre & 74.8\,|\,85.5 & 24.9\,|\,42.3 &  41.5\,|\,59.4 & 40.4\,|\,55.9 \\
 & 0.1\%-Pre & 81.2\,|\,90.7 & 30.1\,|\,46.5 & 41.6\,|\,57.3 & 39.6\,|\,53.9 \\
 & 100\%-Pre & 92.9\,|\,97.6 & 45.9\,|\,62.3 & 55.6\,|\,69.2 & 53.7\,|\,66.4 \\
\midrule
\multirow{5}{*}{7B} & Base (no SFT) & \phantom{0}4.2\,|\,16.3 & \phantom{0}0.2\,|\,\phantom{0}0.8 &  10.1\,|\,18.3 & 12.6\,|\,21.0 \\
 & Base (Relaxed) & 61.8\,|\,84.2 & \phantom{0}4.8\,|\,13.3 & 10.6\,|\,18.8 & 13.3\,|\,21.6 \\
 & 0\%-Pre & 79.5\,|\,89.2 & 27.1\,|\,48.9  & 44.4\,|\,62.2 & 42.9\,|\,57.4 \\
 & 0.1\%-Pre & 88.0\,|\,95.3 & 39.2\,|\,59.1 & 52.3\,|\,66.6 & 50.2\,|\,63.6 \\
 & 100\%-Pre & 93.6\,|\,97.6 & 47.9\,|\,65.6 & 58.7\,|\,71.2 & 56.0\,|\,67.8 \\
\midrule
\multirow{5}{*}{14B} & Base (no SFT) & 13.6\,|\,38.8 & \phantom{0}0.7\,|\,\phantom{0}2.4 &  \phantom{0}5.8\,|\,14.9 & \phantom{0}6.1\,|\,14.8 \\
 & Base (Relaxed) & 56.5\,|\,82.5 & \phantom{0}4.3\,|\,12.3 & \phantom{0}8.8\,|\,18.4 & \phantom{0}8.6\,|\,18.1 \\
 & 0\%-Pre & 79.2\,|\,89.9 & 29.0\,|\,48.5 &  51.0\,|\,65.9 & 48.5\,|\,62.9 \\
 & 0.1\%-Pre & 88.0\,|\,94.9 & 42.2\,|\,59.2  & 53.9\,|\,69.5 & 52.0\,|\,65.0 \\
 & 100\%-Pre & 92.5\,|\,96.8 & 44.8\,|\,60.3 & 60.3\,|\,72.6 & 57.8\,|\,69.6 \\
\bottomrule
\end{tabular}
\caption{Results for the different models over the math and code benchmarks. The 0\%-Pre models show non-negligible results, effectively beating the base models while slightly underperforming their 100\%-Pre counterparts.
The 0.1\%-Pre models narrow the gap in most configurations.
}
\label{tab:merged_results}
\end{table}

We now turn to comparing pre-trained and non-pretrained models when heavily post-trained on specific distributions. If post-training merely unlocks pre-existing capabilities, the non-pretrained models should fail entirely as opposed to their pre-trained counterparts. In contrast, if it functions as a distribution-fitting mechanism, these non-pretrained models should achieve non-trivial performance. 

We report the average over the math competition benchmarks alongside the other benchmarks in \cref{tab:merged_results}, while detailed per-benchmark math results are deferred to \cref{tab:omr_competition} in \cref{app:math_results}.
As expected, the Qwen-Base models, which do not undergo any post-training, exhibit poor performance across all benchmarks under strict-format evaluation. While basic performance slightly increases with model size, the absolute accuracy remains low.
These results are unsurprising given our use of a zero-shot evaluation paradigm, which is generally sub-optimal for base models on competitive benchmarks. For completeness, we also report base-model results using relaxed rules to extract the final answer. Further details are given in~\cref{app:exp_setup}. \looseness=-1

Applying the conventional SFT paradigm to these base models using OMR and \mbox{OCR-II}~(100\%-Pre-Math/Code) yields substantial performance increases across all benchmarks and scales. E.g., the 7B 100\%-Pre-Math model surges to 93.6\% on MATH-500 and 47.9\% on the competition math average. This significant boost is anticipated, aligning with established trends in current LLM training methodologies~\citep{cwm,omr,ocr}.

However, when considering the from-scratch models (0\%-Pre-Math/Code), we observe that despite starting with randomly initialized weights, they achieve surprisingly competitive results by relying entirely on the in-domain SFT distribution. Across all sizes, the models trained from scratch significantly outperform the standard base models, lagging behind their fully pre-trained counterparts by a margin of only 8\% to 20\%. For instance, the 14B 0\%-Pre-Code model achieves 51\% pass@1 on LCB V5, compared to the base model's 6/8.8\% under strict/relaxed evaluation, and to the 60\% of the 100\%-Pre-Code model with the same size. This is also evident when examining the math models, as can be seen for the 1.5B 0\%-Pre-Math model that lags behind its fully pre-trained counterpart by only 10--20\%, while achieving impressive results (74/25\% at MATH-500/competition-math-average). These gaps mirror the GPT-1 and BERT era \citep{gpt1,bert}, where pre-trained models outperformed their supervised-only counterparts by similar margins.

We provide thinking trajectories and answers for the 0\%-Pre models in \cref{app:examples}. As shown, these non-pretrained models generate coherent, highly accurate reasoning chains which often produce correct answers for highly tracked benchmarks. Their ability to succeed in competitive programming and competition math solely through the SFT phase highlights how heavily post-training dominates task-specific performance.

\paragraph{Cross-domain generalization.} A key question that arises is whether the post-training setup enables cross-domain generalization. E.g., whether the intensive post-training on math might yield superior performance on code. To test this hypothesis, we also conduct a cross-domain evaluation on the examined reasoning benchmarks. The results are presented in \cref{tab:cross_merged} in \cref{app:cross_domain}. As expected, having seen no data from the alternate distribution, the 0\%-Pre models demonstrate no cross-domain generalization and perform poorly. Importantly, we observe similar failures in the fully pre-trained models~(100\%-Pre), indicating that optimizing for a specific distribution in isolation is insufficient for reasoning generalization.\footnote{We note that standard post-training approaches often combine datasets from multiple domains. Studying the generalization effect in such setups is an important topic for future work. }

\paragraph{Minimal pre-training setup.}
To further understand the role of the pre-training phase for specific desired behaviors, we add for each model scale a \textit{short pre-trained} version, where we first pre-train the model using general data on a fairly low amount of tokens (10B tokens, less than $\nicefrac{1}{1,000}$ of standard pre-training tokens compared to \cite{qwen_25}), then perform the SFT phase as for the other models. These models are trained on a limited sample of the DCLM dataset~\citep{dclm}, before the SFT phase. See more details in~\cref{app:exp_setup}.

The results for the minimal pre-training phase (0.1\%-Pre-Math/Code) are presented in \cref{tab:merged_results}~(with math competition results in \cref{app:math_results}). Across the majority of models~(excluding 1.5B 0.1\%-Pre and 3B 0.1\%-Pre-Code), this minimal initialization phase successfully closes 25\% to 84\% of the gap between from-scratch and full pre-training. 0.1\%-Pre-Code at the 7B scale, for example, reaches 50.2\% pass@1 on LCB V6, halving the distance between the from-scratch model (42.9\%) and the fully pre-trained one (56.0\%). We hypothesize that these gains stem from improved weight initialization and basic natural language acquisition, rather than exposure to domain-specific data (which is not explicitly included in the short pre-training phase). This suggests that the major part of the  performance gap between pre-trained models and their non-pretrained counterparts is largely structural, serving as a better starting point for the SFT phase instead of adding specific knowledge.

\begin{figure*}[t!]
     \centering
     \begin{subfigure}[b]{0.24\textwidth}
         \centering
         \includegraphics[trim={0.3cm 0.25cm 0.2cm 0.2cm},clip,width=\textwidth]{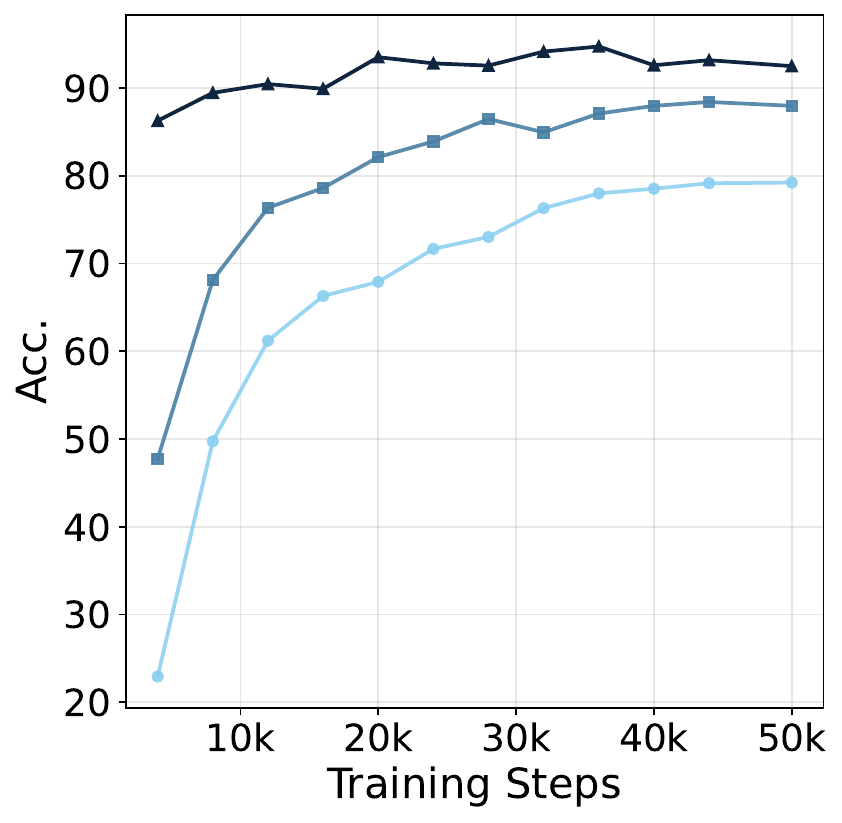}
         \caption{Math-500 \label{fig:14b_math}}
     \end{subfigure}
     \hfill
     \begin{subfigure}[b]{0.24\textwidth}
         \centering
         \includegraphics[trim={0.3cm 0.25cm 0.2cm 0.2cm},clip,width=\textwidth]{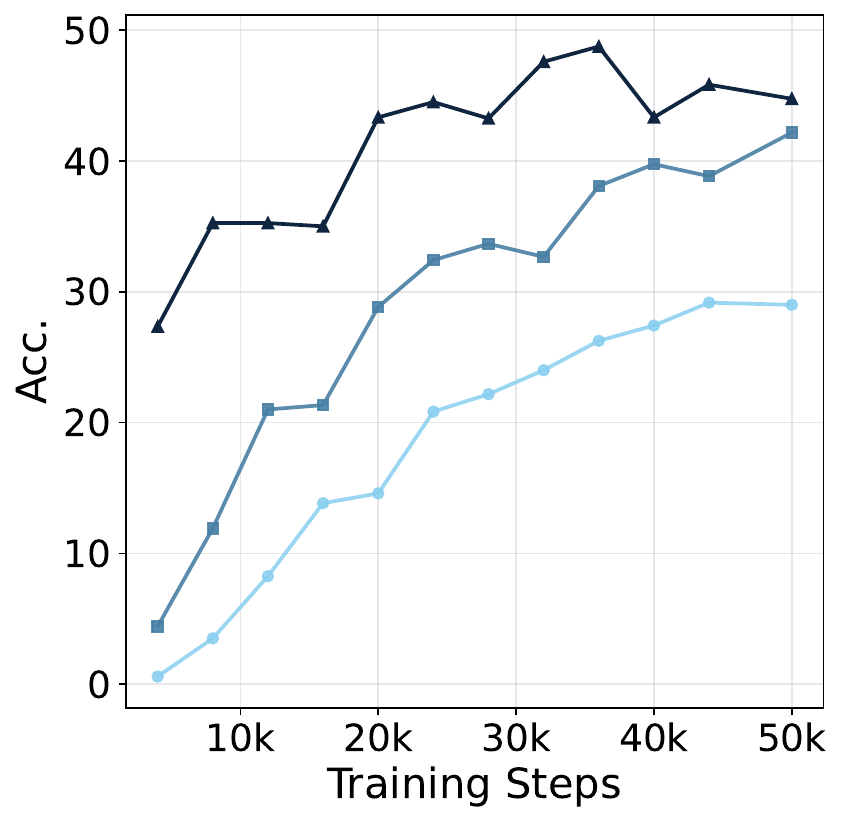}
         \caption{Competition Math \label{fig:14b_avg}}
     \end{subfigure}
     \hfill
    \begin{subfigure}[b]{0.24\textwidth}
         \centering
         \includegraphics[trim={0.3cm 0.25cm 0.2cm 0.2cm},clip,width=\textwidth]{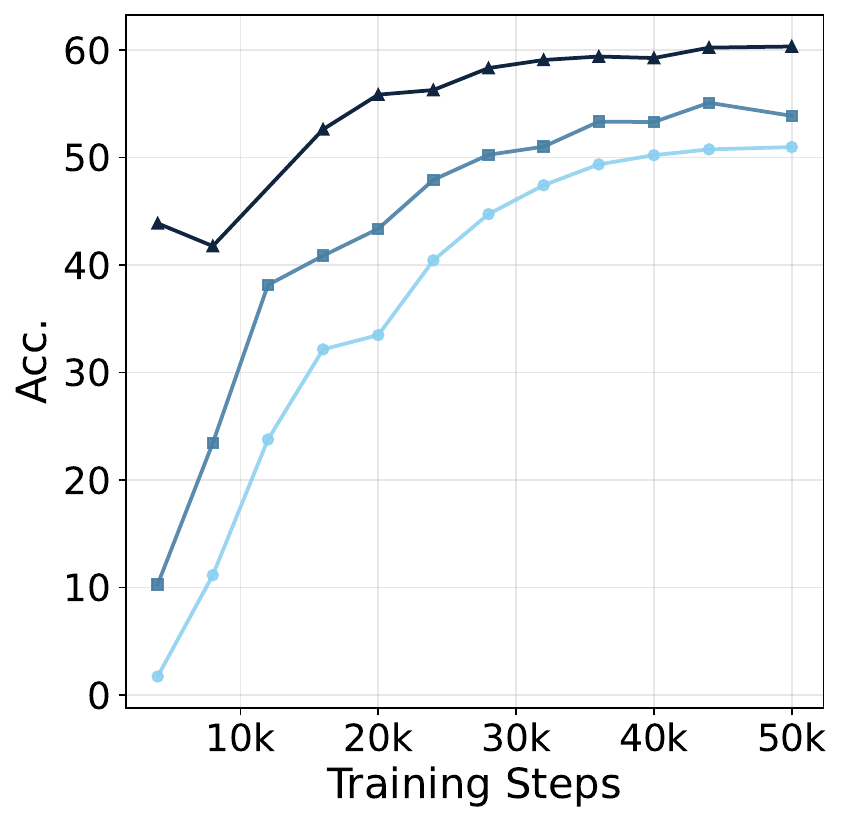}
         \caption{LCB V5 \label{fig:14b_lcb_v5}}
     \end{subfigure}
     \hfill
     \begin{subfigure}[b]{0.24\textwidth}
         \centering
         \includegraphics[trim={0.3cm 0.25cm 0.2cm 0.2cm},clip,width=\textwidth]{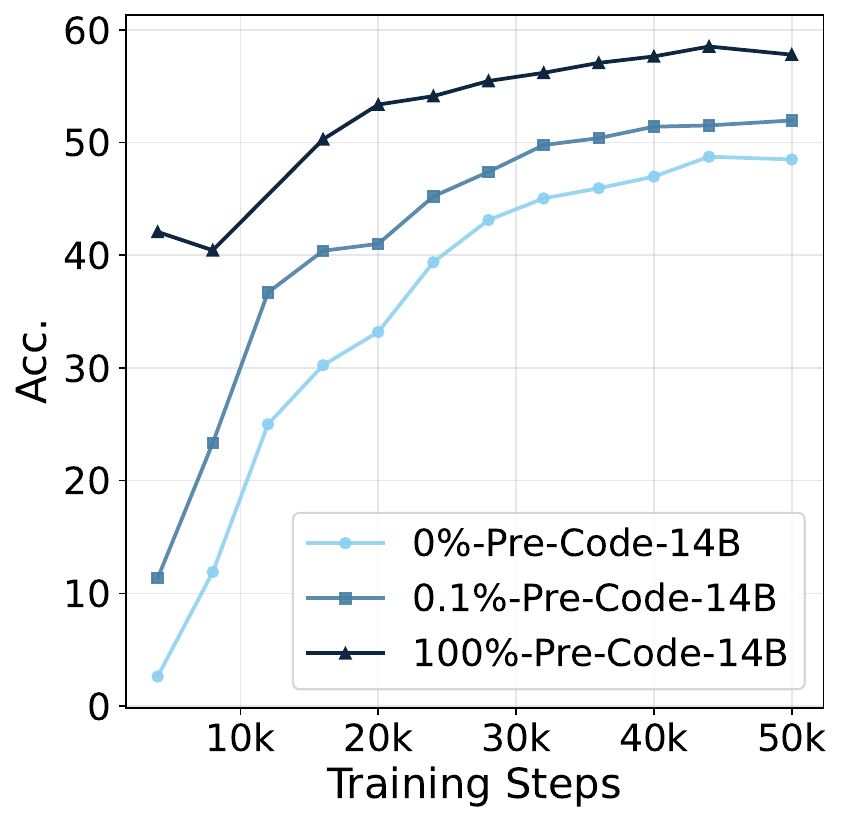}
         \caption{LCB v6 \label{fig:14b_lcb_v6}}
     \end{subfigure}
     \caption{Evaluation curves during SFT using different amounts of pre-training backbones for the 14B scale. The  0/0.1\%-Pre models show rapid improvement over the first 20k steps.\label{fig:14b}}
\end{figure*}

\paragraph{Performance during training.} Finally, we explore the performance of the different models over the course of training. \cref{fig:14b} presents the performance of the 14B model as training progresses, while \cref{fig:7b,fig:3b,fig:1b} in \cref{app:small_model_results} present the curves of other models (\cref{app:math_curves} presents detailed math competition curves). As can be seen, the 100\%-Pre-Math/Code models adapt to the different distributions relatively fast. Although they continue to improve later in training, they achieve the majority of their performance gains halfway through the SFT phase. Conversely, the 0\%-Pre and 0.1\%-Pre models exhibit weaker performance during the first few thousand training steps (due to a lack of prior knowledge), but quickly catch up, showing rapid improvement over the first 20k steps. While this again confirms the utility of pre-training, it is highly interesting to observe how well and relatively fast models can perform in-distribution relying solely on supervised data.

%% file: 04_discussion.tex
\section{Learning how to Learn}
\label{sec:disc}

As discussed in \cref{sec:story}, the evolution of LLMs has transitioned from pre-trained backbones used for task-specific SFT to a massive post-training paradigm that aligns desired behaviors into the model's distribution. Yet, this current pipeline remains fundamentally reactive: the post-training process is inherently restricted to specific use cases and behaviors that are \textit{already known}, and for which large-scale post-training can be performed. 
These models demonstrate robust generalization within their established training distributions. However, adapting them to entirely novel behaviors necessitates full re-(post-)training cycles, a limitation that contrasts with the  promise of few-shot learning~\citep{gpt3}.

This limitation drives the current trend of iterative model releases built upon the same pre-trained backbone, distinguished only by enhanced post-training setups that incorporate newly desired capabilities (e.g., the incremental progressions of GPT-5.x or Claude-4.x). Under the current paradigm, acquiring a novel capability demands a costly, ground-up re-post-training effort: curating new SFT datasets, engineering new RL environments, and executing continuous model updates just to bridge the newly identified capability gap. 
Proponents of the current paradigm may argue that scaling post-training data and compute can lead to achieving broader capabilities. Yet, this relies on the assumption that scaling predefined data and environments can efficiently keep pace with the ever-evolving demands placed on LLM systems. Therefore, we posit that the field must advance beyond these static, predefined post-training procedures toward a new evolutionary stage.

We argue for the development of significantly more robust systems. In our vision, models must inherently \textbf{\textit{``learn how to learn''}}. Under this proposed paradigm, when an LLM system encounters a problem it cannot solve to the required degree of accuracy, it should autonomously adapt and evolve its capabilities to overcome that limitation.

One approach to realizing this vision is through autonomous, self-adapting LLM agents. We envision an agentic system capable of dynamically modifying its own prompts, underlying code, and even its model weights. When presented with a novel task where superficial adjustments (such as prompt engineering or modifying code execution) are insufficient, the system could autonomously replicate the LLM post-training pipeline. Specifically, it could search for and curate SFT data, engineer custom RL environments, and subsequently retrain its internal models. This method can include attempting a solution, while iteratively learning through any failures.
The system would then evaluate the updated model's performance on the target task, iterating on this process as necessary. Furthermore, under resource constraints, such a system could intelligently opt to train a smaller, highly specialized model to serve as a tool for future instances of the new task.

Of course, the acquisition of the above meta-learning capabilities must be integrated as a core objective within modern training paradigms. Just as traditional pre-training serves as the foundational phase for general knowledge and linguistic skills, this proposed meta-learning phase acts as the mechanism for dynamic, post-deployment task acquisition (similar to the promise of few-shot learning). 
Ultimately, this establishes a new level of machine intelligence defined by the ability to ``learn how to learn''.

Crucially, this foundational training would ideally only be required once. Once a model acquires the capacity to self-evolve, it will be able to theoretically improve indefinitely, limited only by computational resources. Consequently, this paradigm will relieve researchers from continuously engineering and maintaining post-training procedures by delegating the control of ongoing adaptation directly to the models themselves.

We note that our suggested paradigm idea is not limited to the specific mechanism described above, and can have several forms and generalizability \citep{bandel2026position}. Various approaches can successfully enable systems to learn how to solve unknown tasks \citep{autoflow,recursive_lms,zhang2025darwin,wang2026huxleygodel,evolveR,survey_1,survey_2,hyper,weng2026group, xia2026skillrl}. Achieving this self-evolving capacity may also require moving beyond standard Transformer backbones to alternative architectures that yield highly effective in-context learning for novel tasks without massive supervised learning \citep{zhang2025memory}.
The unifying idea for optimizing all such systems is the shared principle of ``learning how to learn''. Different approaches to this objective will naturally result in varying systems, which will result in a robust ecosystem of self-improving models.

%% file: 06_conclusion.tex
\section{Conclusion}
\label{sec:conc}

In this position paper, through a historical overview of LLMs, we argued that current training paradigms effectively revert to the ``pre-train then fine-tune'' methodology of the BERT era. Under this approach, models are explicitly tailored to desired behaviors and specific evaluation benchmarks. We empirically demonstrated this by showing that models post-trained from scratch yield highly non-trivial performance, suggesting that current post-training methodologies function primarily as a distribution-fitting mechanism. Finally, we argued that prevailing post-training approaches are fundamentally reactive. To achieve more generalized capabilities, we advocate for a shift toward meta-learning paradigms, where models ``learn how to learn''.

%% file: 92_exp_setup.tex
\section{Detailed Experimental Setup}
\label{app:exp_setup}

\paragraph{Evaluation benchmarks.} The MATH-500 comprises 500 high school math questions \citep{math500}. The competition math benchmarks, AIME 24, AIME 25, HMMT February 25, and HMMT November 25, taken from the MathArena leaderboard \citep{matharena}. Each of them contains 30 competition level math problems. The LCB coding benchmark is built from  competitive programming questions, where the V5 splits consists of 327 problems and the V6 split has 454 problems \citep{lcb}.

\paragraph{SFT datasets.} The problems of the OMR dataset were collected from AoPS forums,\footnote{ \url{artofproblemsolving.com/community}} while the solutions were generated using DeepSeek-R1 and Qwen-QwQ \citep{r1, qwq}. As for OCR-II, it aggregates programming challenges from TACO \citep{taco}, APPS \citep{apps}, CodeContests \citep{codecontest}, and CodeForces \citep{codeforces}, pairing them with solutions that were generated by DeepSeek-R1 and subsequently verified by Qwen-QwQ.

\paragraph{Short pre-training details.} The minimal pre-training phase is based on the DCLM dataset~\citep{dclm}, which was constructed as a high-quality dataset for pre-training. This phase uses $\sim$10B tokens, which is strictly less than $\nicefrac{1}{1,000}$ of the original pre-training corpus used by Qwen-2.5-Base. The short pre-training phase retains the configuration of the SFT phase below but employs a constant learning rate of 1e-4 and 500 steps for warm-up.

\paragraph{Artifact Licenses.} 
Regarding the evaluation benchmarks, MATH-500 is licensed under Apache~2.0~\citep{math500}, while MathArena and LCB are licensed under the MIT license~\citep{matharena, lcb}. The OMR and OCR training datasets are released under the Creative Commons Attribution 4.0 International License (CC BY 4.0; \citealp{omr,ocr2}), and the DCLM dataset is distributed under the MIT license \citep{dclm}. Finally, the Qwen models are licensed under Apache~2.0~\citep{qwen_25}.

\paragraph{Training and evaluation details.} We use the AdamW optimizer \citep{adamw} with a peak learning rate of 3e-4, a cosine decay scheduler, and a 2,000-step linear warmup. All experiments were conducted using NVIDIA H100 GPU nodes, with total computational usage amounting to about 4,000 GPU hours.
For evaluation, we use zero-shot prompt style, which evaluates the ability of models also to follow instructions. We detail our math and code prompts templates below.

\begin{figure}[h]
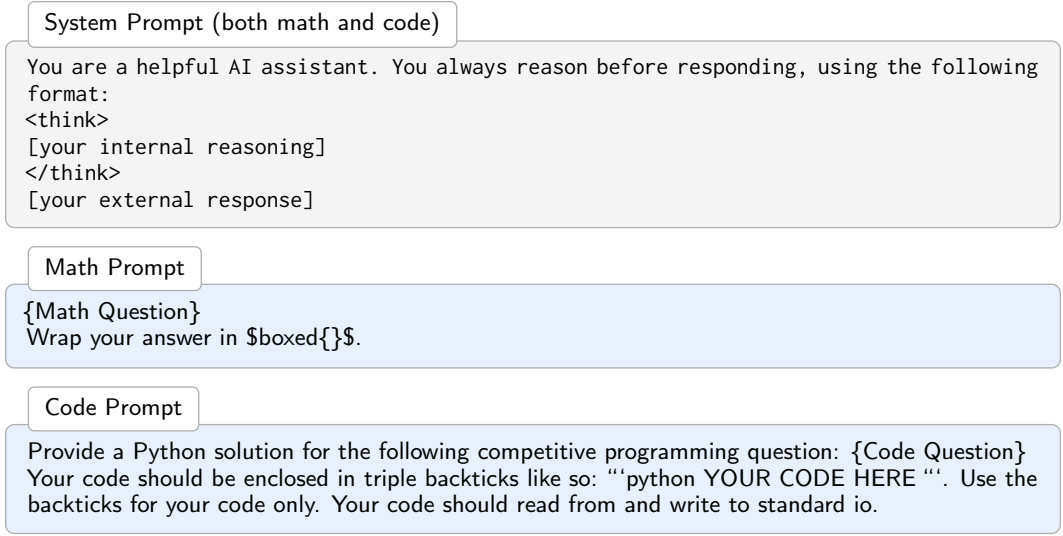

    \centering
    \begin{promptbox}[title=System Prompt (both math and code)]{systemgray}
    \small \ttfamily
    You are a helpful AI assistant. You always reason before responding, using the following format:
    
    <think>
    
    [your internal reasoning]
    
    </think>
    
    [your external response]
    \end{promptbox}
    
    \vspace{-1.5mm}
    
    \begin{promptbox}[title=Math Prompt]{userblue}
    \small \sffamily
    \{Math Question\}
    
    Wrap your answer in \$boxed\{\}\$.
    
    \end{promptbox}

    \vspace{-1.5mm}

    \begin{promptbox}[title=Code Prompt]{userblue}
    \small \sffamily
    Provide a Python solution for the following competitive programming question: \{Code Question\}
    
    Your code should be enclosed in triple backticks like so: ```python YOUR CODE HERE ```. Use the backticks for your code only. Your code should read from and write to standard io.
    \end{promptbox}
    \caption{Prompt templates for the math and code questions.}
    \label{fig:prompt_template}
\end{figure}

\paragraph{Relaxed (Non-strict) Answer Extraction.}

For the relaxed evaluation protocol, we search the complete output for the final answer without requiring a boxed math answer or a Python code fence.

For math benchmarks, rule-based scanners collect candidates from boxed expressions, answer or conclusion statements, and ending lines. Then they select the last candidate classified as a valid answer and compare it with the reference. For code benchmarks, rules collect syntax-valid programs and select a final candidate before running tests. Selection prefers labeled solutions and matching function entrypoints, otherwise using the last complete fence or the longest valid span.

%% file: 91_appendix_full_math_results.tex
\section{Full Competition Math Results}
\label{app:math_results}

We present the benchmark-specific results for the competition math benchmarks: AIME-24, AIME-25, HMMT February 25 and HMMT November 25 in \cref{tab:omr_competition}.

\begin{table}[h]
\centering
\tiny
\setlength{\tabcolsep}{4.5pt}
\begin{tabular}{llcccc}
\toprule
\textbf{Size} & \textbf{Model} & \textbf{AIME'24} & \textbf{AIME'25} & \textbf{HMMT Feb} & \textbf{HMMT Nov} \\
 &  & {pass@1\,|\,5} & {pass@1\,|\,5} & {pass@1\,|\,5} & {pass@1\,|\,5} \\
\midrule
\multirow{5}{*}{1.5B} & Base (no SFT) & \phantom{0}0.0\,|\,\phantom{0}0.0 & \phantom{0}0.0\,|\,\phantom{0}0.0 & \phantom{0}0.0\,|\,\phantom{0}0.0 & \phantom{0}0.0\,|\,\phantom{0}0.0 \\
 & Base (Relaxed) & \phantom{0}0.3\,|\,\phantom{0}1.7 & \phantom{0}0.0\,|\,\phantom{0}0.0 & \phantom{0}0.0\,|\,\phantom{0}0.0 & \phantom{0}0.0\,|\,\phantom{0}0.0 \\
 & 0\%-Pre-Math & 32.0\,|\,57.6 & 30.7\,|\,50.1 & 16.7\,|\,32.7 & 19.3\,|\,33.2 \\
 & 0.1\%-Pre-Math & 26.7\,|\,51.1 & 21.7\,|\,34.5 & 10.7\,|\,24.4 & 16.0\,|\,26.6 \\
 & 100\%-Pre-Math & 56.0\,|\,77.0 & 38.0\,|\,61.9 & 26.0\,|\,52.8 & 35.3\,|\,54.0 \\
\midrule
\multirow{5}{*}{3B} & Base (no SFT) & \phantom{0}1.3\,|\,\phantom{0}5.2 & \phantom{0}0.0\,|\,\phantom{0}0.0 & \phantom{0}0.3\,|\,\phantom{0}1.7 & \phantom{0}0.3\,|\,\phantom{0}1.7 \\
 & Base (Relaxed) & \phantom{0}1.3\,|\,\phantom{0}4.7 & \phantom{0}0.7\,|\,\phantom{0}3.3 & \phantom{0}0.3\,|\,\phantom{0}1.7 & \phantom{0}0.3\,|\,\phantom{0}1.7 \\
 & 0\%-Pre-Math & 40.7\,|\,60.9 & 26.7\,|\,46.4 & 10.7\,|\,20.0 & 21.7\,|\,42.0 \\
 & 0.1\%-Pre-Math & 44.0\,|\,68.1 & 35.7\,|\,51.3 & 17.0\,|\,28.9 & 23.7\,|\,37.8 \\
 & 100\%-Pre-Math & 62.7\,|\,79.1 & 47.0\,|\,64.1 & 33.0\,|\,50.3 & 41.0\,|\,55.6 \\
\midrule
\multirow{5}{*}{7B} & Base (no SFT) & \phantom{0}0.3\,|\,\phantom{0}1.7 & \phantom{0}0.0\,|\,\phantom{0}0.0 & \phantom{0}0.0\,|\,\phantom{0}0.0 & \phantom{0}0.3\,|\,\phantom{0}1.7 \\
 & Base (Relaxed) & \phantom{0}9.3\,|\,20.1 & \phantom{0}5.7\,|\,17.9 & \phantom{0}0.0\,|\,\phantom{0}0.0 & \phantom{0}4.3\,|\,15.1 \\
 & 0\%-Pre-Math & 42.7\,|\,69.4 & 27.7\,|\,51.6 & 14.7\,|\,32.8 & 23.3\,|\,42.0 \\
 & 0.1\%-Pre-Math & 50.7\,|\,73.7 & 44.3\,|\,63.0 & 28.7\,|\,49.5 & 33.0\,|\,50.3 \\
 & 100\%-Pre-Math & 67.0\,|\,79.9 & 48.0\,|\,70.3 & 32.3\,|\,53.2 & 44.3\,|\,59.0 \\
\midrule
\multirow{5}{*}{14B} & Base (no SFT) & \phantom{0}1.3\,|\,\phantom{0}4.7 & \phantom{0}0.0\,|\,\phantom{0}0.0 & \phantom{0}0.3\,|\,\phantom{0}1.7 & \phantom{0}1.0\,|\,\phantom{0}3.1 \\
 & Base (Relaxed) & \phantom{0}7.3\,|\,23.6 & \phantom{0}5.7\,|\,12.6 & \phantom{0}2.0\,|\,\phantom{0}8.1 & \phantom{0}2.3\,|\,\phantom{0}5.0 \\
 & 0\%-Pre-Math & 41.7\,|\,64.3 & 30.3\,|\,46.6 & 18.7\,|\,38.1 & 25.3\,|\,45.2 \\
 & 0.1\%-Pre-Math & 57.7\,|\,72.3 & 44.7\,|\,65.8 & 32.3\,|\,54.1 & 34.0\,|\,44.6 \\
 & 100\%-Pre-Math & 60.3\,|\,75.1 & 47.0\,|\,61.8 & 31.3\,|\,49.1 & 40.3\,|\,55.3 \\
\bottomrule
\end{tabular}
\caption{Results for the math models over the different math competition benchmarks.}
\label{tab:omr_competition}
\end{table}

%% file: 94_examples.tex
\section{Examples of Generated Outputs}
\label{app:examples}

\cref{fig:aime_25_ex} presents an example response from the 0\%-Pre-Math (14B) model for an AIME 25 question, while \cref{fig:lcb_v6_ex} demonstrates the 0\%-Pre-Code (7B) model's response to an LCB V6 question. As can be seen, the reasoning trajectory is coherent and understandable, although the model was trained without any prior linguistic knowledge.

\begin{figure*}[h]
    \centering

    \begin{promptbox}[title=Math Question]{userblue}
    \small \ttfamily
    Find the sum of all integer bases $b > 9$ for which $17_b$ is a divisor of $97_b$.
    Wrap your answer in \$boxed\{\}\$
    \end{promptbox}
    
    \vspace{-1.5mm} 
  
    \begin{promptbox}[title=0\%-Pre-Math (14B) Response]{modelpurple}
    \small \sffamily
    <think> Okay, let's see. I need to find all integers b greater than 9 such that $17_b$ divides $97_b$. Then sum those b's. Hmm.
    First, let's convert both numbers from base b to base 10 to work with them more easily.
    The number 17 in base b is equal to 1*b + 7 in decimal. Similarly, 97 in base b is 9*b + 7 in decimal.
    So, we need (1*b + 7) divides (9*b + 7). Let me write that as an equation:
    (9b + 7) divided by (b + 7) should be an integer.
    Let me compute (9b + 7) ÷ (b + 7). Maybe perform the division to see what the quotient and remainder are.
    Divide 9b + 7 by b + 7. Let's see:
    How many times does (b + 7) go into 9b + 7? Let's do polynomial division.
    Divide 9b by b, which is 9. Multiply (b + 7) by 9: 9b + 63. Subtract that from 9b + 7: (9b + 7) - (9b + 63) = -56. So the division gives 9 with a remainder of -56. Therefore, (9b + 7) = 9*(b + 7) - 56.
    So, (9b + 7)/(b + 7) = 9 - 56/(b + 7). For this to be an integer, 56/(b + 7) must be an integer. Therefore, (b + 7) must be a divisor of 56.
    But since b > 9, then b + 7 > 16. So, we need to find all divisors of 56 that are greater than 16, then subtract 7 to get b.First, list all positive divisors of 56. 56 factors into $2^3 * 7^1$. So the divisors are 1, 2, 4, 7, 8, 14, 28, 56.From these, the divisors greater than 16 are 28 and 56.
    Therefore, $b + 7 = 28 \implies b = 21$, and $b + 7 = 56 \implies b = 49$.
    So the possible values of b are 21 and 49. Summing them gives 21 + 49 = 70.
    Wait, but let me check. Let's verify each b.
    First, b = 21. Then $17_{21}$ is $1*21 + 7 = 28$. $97_{21}$ is $9*21 + 7 = 189 + 7 = 196$. Check if 28 divides 196. $196 ÷ 28 = 7$, which is an integer. So that's good.
    Next, b = 49. $17_{49}$ is 1*49 + 7 = 56. $97_{49}$ is 9*49 + 7 = 441 + 7 = 448. 448 ÷ 56 = 8, which is also an integer. So that works too.
    Are there any other divisors of 56 greater than 16? Let's check the list again. The divisors are 1, 2, 4, 7, 8, 14, 28, 56. So after 14 comes 28 and 56. So only 28 and 56. So yes, only 21 and 49. Therefore, the sum is 70.
    I think that's it. Let me make sure there's no mistake in the logic. The key step was realizing that (9b +7) divided by (b +7) must be integer, which requires that the remainder is zero. The remainder was -56, so (b +7) must divide 56. Then considering the divisors of 56 greater than 16. That seems correct. So the answer should be 21 + 49 = 70.</think> 
    
    To solve the problem of finding the sum of all integer bases $ b > 9 $ for which $ 17_b $ divides $ 97_b $, we proceed as follows:
    
    1. **Convert the numbers from base $ b $ to base 10:**   - The number $ 17_b $ in base 10 is $ 1 \\cdot b + 7 = b + 7 $.
    
    - The number $ 97_b $ in base 10 is $ 9 \\cdot b + 7 = 9b + 7 $.
    
    2. **Set up the divisibility condition:**   - We need $ b + 7 $ to divide $ 9b + 7 $. This means $ \\frac{9b + 7}{b + 7} $ must be an integer.
    
    3. **Perform the division:**   - Divide $ 9b + 7 $ by $ b + 7 $:     $     \\frac{9b + 7}{b + 7} = 9 - \\frac{56}{b + 7}     $   - For this to be an integer, $ \\frac{56}{b + 7} $ must be an integer. Therefore, $ b + 7 $ must be a divisor of 56.
    
    4. **Identify the divisors of 56:**   - The divisors of 56 are: 1, 2, 4, 7, 8, 14, 28, 56.   - Since $ b > 9 $, we need $ b + 7 > 16 $. Thus, the relevant divisors are 28 and 56.
    
    5. **Determine the corresponding values of $ b $:**   - If $ b + 7 = 28 $, then $ b = 21 $.   - If $ b + 7 = 56 $, then $ b = 49 $.
    
    6. **Verify the solutions:**   - For $ b = 21 $:     $     17_{21} = 1 \cdot 21 + 7 = 28     $     $     97_{21} = 9 \cdot 21 + 7 = 189 + 7 = 196     $     $     196 \div 28 = 7 \quad (\text{an integer})     $   - For $ b = 49 $:     $     17_{49} = 1 \cdot 49 + 7 = 56     $     $     97_{49} = 9 \cdot 49 + 7 = 441 + 7 = 448     $     $     448 \div 56 = 8 \quad (\text{an integer})     $
    
    7. **Sum the valid values of $ b $:**   $   21 + 49 = 70   $
    
    Thus, the sum of all integer bases $ b > 9 $ for which $ 17_b $ divides $ 97_b $ is $\boxed{70}$.
    \end{promptbox}

    \caption{A response example of 0\%-Pre-Math (14B) on an example from AIME 25.}
    \label{fig:aime_25_ex}
\end{figure*}

\begin{figure*}[h]
    \centering

    \begin{promptbox}[title=Code Question]{userblue}
    \small \ttfamily
    Provide a Python solution for the following competitive programming question: There is a stack of N cards, and the i-th card from the top has an integer A\_i written on it. You take K cards from the bottom of the stack and place them on top of the stack, maintaining their order. Print the integers written on the cards from top to bottom after the operation.
    Input
    The input is given from Standard Input in the following format:N K
    A\_1 A\_2 $\ldots$ A\_N
    
    Output
    
    Let B\_i be the integer written on the i-th card from the top of the stack after the operation. Print B\_1,B\_2,\\ldots,B\_N in this order, separated by spaces.
    
    Constraints
    $1 \leq K < N \leq 100$ ; $ 1 \leq A_i \leq 100$- All input values are integers.

    \{Example IO pairs are omitted\}
    
    Your code should be enclosed in triple backticks like so: ```python YOUR CODE HERE ```. Use the backticks for your code only. Your code should read from and write to standard io.
    \end{promptbox}
    
    \vspace{-1.5mm} 
  
    \begin{promptbox}[title=0\%-Pre-Code (7B) Response]{modelpurple}
    \small \sffamily
    <think> Okay, let's see. The problem is about moving K cards from the bottom of a stack to the top, keeping their order. So, for example, if the original stack is [1,2,3,4,5] (with 1 being the top), and K is 3, then we take the bottom three cards (3,4,5) and put them on top. So the new stack becomes [3,4,5,1,2]. Then we need to output them top to bottom. Hmm. So the input is N and K, then the list A which is the initial stack from top to bottom. Wait, wait. Wait, the problem says \"the i-th card from the top has A\_i written on it.\" So the first element in the input list is the top card, and the last is the bottom. So the initial stack is A\_1 (top), A\_2, ..., A\_N (bottom). So when we take K cards from the bottom, which would be the last K elements of the list. Then we move those to the top. So the new stack is [A\_{N-K+1}, A\_{N-K+2}, ..., A\_N] followed by [A\_1, A\_2, ..., A\_{N-K}]. Wait, no. Wait, the original stack is A\_1 (top), A\_2, ..., A\_N (bottom). So the bottom K cards are A\_{N-K+1}, A\_{N-K+2}, ..., A\_N. Because if N is 5 and K is 3, then the bottom 3 cards are positions 3,4,5 (since 5-3+1=3). Wait, no. Wait, if the stack is 1 (top), 2, 3,4,5 (bottom), then the bottom three cards are 3,4,5. So the indices in the list would be [2], [3], [4] (assuming zero-based). Wait, the list is given as A\_1 to A\_N. So for the sample input 1, the list is [1,2,3,4,5]. The bottom three cards are 3,4,5. So their indices in the list are 2,3,4 (since Python is zero-based). So to get the bottom K elements, we need to take the last K elements of the list. So the process is: take the last K elements (from the bottom) and put them on top. So the new list is (last K elements) + (first N-K elements). Then, when we output, we print each element in order, which is the new top to bottom. So for sample input 1, the original list is [1,2,3,4,5]. Take last 3 elements [3,4,5], then add the first 2 elements [1,2]. So new list is [3,4,5,1,2]. Then output as space-separated. So the steps are: 1. Read N and K from input. 2. Read the list A as integers. 3. Split the list into two parts: the last K elements (bottom K) and the rest. 4. The new stack is the concatenation of the last K elements and the first N-K elements. 5. Print the new stack elements in order, separated by spaces. So in code: Read N, K = map(int, input().split()) A = list(map(int, input().split())) new\_stack = A[-K:] + A[:-K] print(' '.join(map(str, new\_stack))). Wait, but wait. Let's test with sample input 1. N=5, K=3. A is [1,2,3,4,5]. A[-3:] is [3,4,5], A[:-3] is [1,2]. So concatenating gives [3,4,5,1,2]. Which is correct. Sample input 2: N=6, K=2. A is [1,2,1,2,1,2]. The last two elements are [1,2], the first 4 elements are [1,2,1,2]. So new stack is [1,2] + [1,2,1,2] → [1,2,1,2,1,2]. So output is same as input, which matches the sample output. So the code seems correct. But wait, the problem says that after moving the K cards, the new stack's top is the first element. So the code's approach of taking the last K elements and putting them first (as the new top) is correct. Yes. So the code should work. </think>
    
    ```python
    
    \# Read input
    
    N, K = map(int, input().split())
    
    A = list(map(int, input().split()))
    
    \# Calculate the new stack
    
    new\_stack = A[-K:] + A[:-K]
    
    \# Output the result
    
    print(' '.join(map(str, new\_stack)))
    
    ```
    
    \end{promptbox}

    \caption{A response example of 0\%-Pre-Code (7B) on an example from LCB V6.}
    \label{fig:lcb_v6_ex}
\end{figure*}

%% file: 93_app_cross_domain.tex
\section{Cross-Domain Results}
\label{app:cross_domain}

\begin{table*}[h!]
\centering
\tiny
\setlength{\tabcolsep}{4.5pt}
\begin{tabular}{l||lcc||lcc}
\toprule
 & \multicolumn{3}{c||}{\multirow{2}{*}{\small\textbf{Code models on math}}} & \multicolumn{3}{c}{\multirow{2}{*}{\small\textbf{Math models on code}}} \\
 & \multicolumn{3}{c||}{} & \multicolumn{3}{c}{} \\
\textbf{Size} & \textbf{Model} & \textbf{MATH-500} & \textbf{Comp.\ Math} & \textbf{Model} & \textbf{LCB V5} & \textbf{LCB V6} \\
 & & {pass@1\,|\,5} & {pass@1\,|\,5} & & {pass@1\,|\,5} & {pass@1\,|\,5} \\
\midrule
\multirow{2}{*}{1.5B} & 0\%-Pre-Code & \phantom{0}0.0\,|\,\phantom{0}0.0 & \phantom{0}0.0\,|\,\phantom{0}0.0 & 0\%-Pre-Math & \phantom{0}0.0\,|\,\phantom{0}0.0 & \phantom{0}0.0\,|\,\phantom{0}0.0 \\
 & 100\%-Pre-Code & \phantom{0}0.0\,|\,\phantom{0}0.0 & \phantom{0}0.0\,|\,\phantom{0}0.0 & 100\%-Pre-Math & \phantom{0}0.0\,|\,\phantom{0}0.0 & \phantom{0}0.0\,|\,\phantom{0}0.2 \\
\midrule
\multirow{2}{*}{3B} & 0\%-Pre-Code & \phantom{0}0.0\,|\,\phantom{0}0.0 & \phantom{0}0.0\,|\,\phantom{0}0.0 & 0\%-Pre-Math & \phantom{0}0.0\,|\,\phantom{0}0.0 & \phantom{0}0.0\,|\,\phantom{0}0.0 \\
 & 100\%-Pre-Code & \phantom{0}0.0\,|\,\phantom{0}0.0 & \phantom{0}0.0\,|\,\phantom{0}0.0 & 100\%-Pre-Math & \phantom{0}0.1\,|\,\phantom{0}0.3 & \phantom{0}0.4\,|\,\phantom{0}1.1 \\
\midrule
\multirow{2}{*}{7B} & 0\%-Pre-Code & \phantom{0}0.0\,|\,\phantom{0}0.0 & \phantom{0}0.0\,|\,\phantom{0}0.0 & 0\%-Pre-Math & \phantom{0}0.0\,|\,\phantom{0}0.0 & \phantom{0}0.0\,|\,\phantom{0}0.0 \\
 & 100\%-Pre-Code & \phantom{0}0.1\,|\,\phantom{0}0.3 & \phantom{0}0.0\,|\,\phantom{0}0.0 & 100\%-Pre-Math & \phantom{0}0.0\,|\,\phantom{0}0.0 & \phantom{0}0.2\,|\,\phantom{0}0.4 \\
\midrule
\multirow{2}{*}{14B} & 0\%-Pre-Code & \phantom{0}0.0\,|\,\phantom{0}0.0 & \phantom{0}0.0\,|\,\phantom{0}0.0 & 0\%-Pre-Math & \phantom{0}0.0\,|\,\phantom{0}0.0 & \phantom{0}0.0\,|\,\phantom{0}0.0 \\
 & 100\%-Pre-Code & \phantom{0}0.1\,|\,\phantom{0}0.7 & \phantom{0}0.0\,|\,\phantom{0}0.0 & 100\%-Pre-Math & \phantom{0}0.0\,|\,\phantom{0}0.2 & \phantom{0}0.0\,|\,\phantom{0}0.0 \\
\bottomrule
\end{tabular}
\caption{Results for code models on math benchmarks, and math models on code benchmarks.}
\label{tab:cross_merged}
\end{table*}

We present the cross-domain results of the different models. \cref{tab:cross_merged} presents both the code models performance over the math benchmarks, and the results of the math models over the LCB benchmark.\footnote{We do not detail the benchmark specific results for the competition math benchmarks as they are all 0.0.} As can be seen, for all cases, models fine-tuned on a specific distribution failed to generalize to another.

%% file: 90_appendix_small_model_results.tex
\section{Performance Curves}
\label{app:small_model_results}

\begin{figure*}[h!]
     \centering
     \begin{subfigure}[b]{0.24\textwidth}
         \centering
         \includegraphics[trim={0.3cm 0.25cm 0.2cm 0.2cm},clip,width=\textwidth]{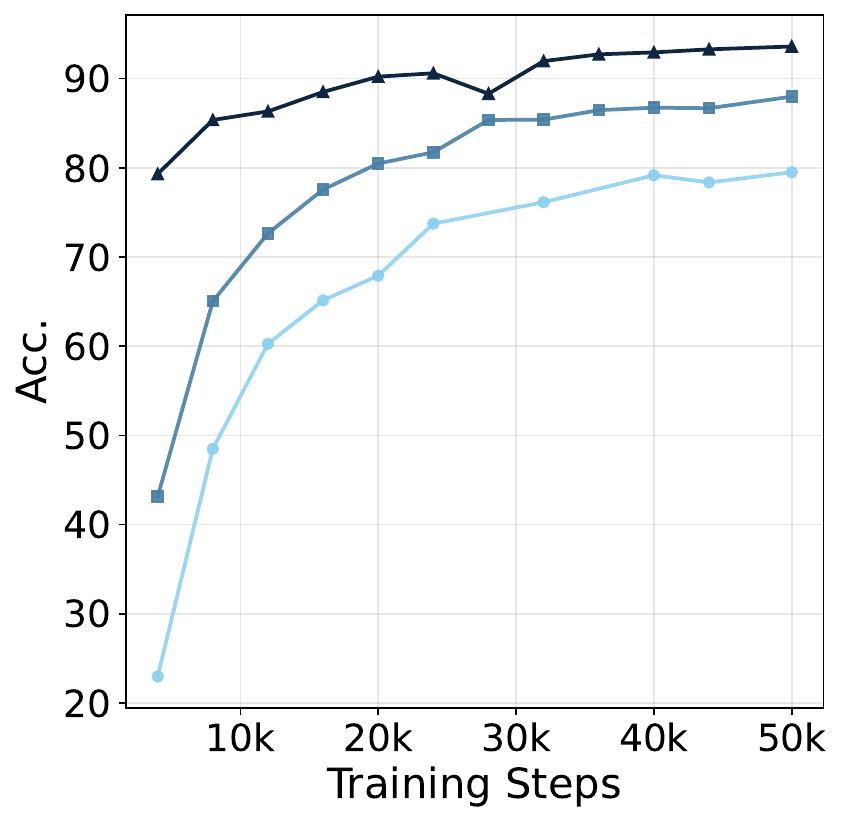}
         \caption{Math-500 \label{fig:7b_math}}
     \end{subfigure}
     \hfill
     \begin{subfigure}[b]{0.24\textwidth}
         \centering
         \includegraphics[trim={0.3cm 0.25cm 0.2cm 0.2cm},clip,width=\textwidth]{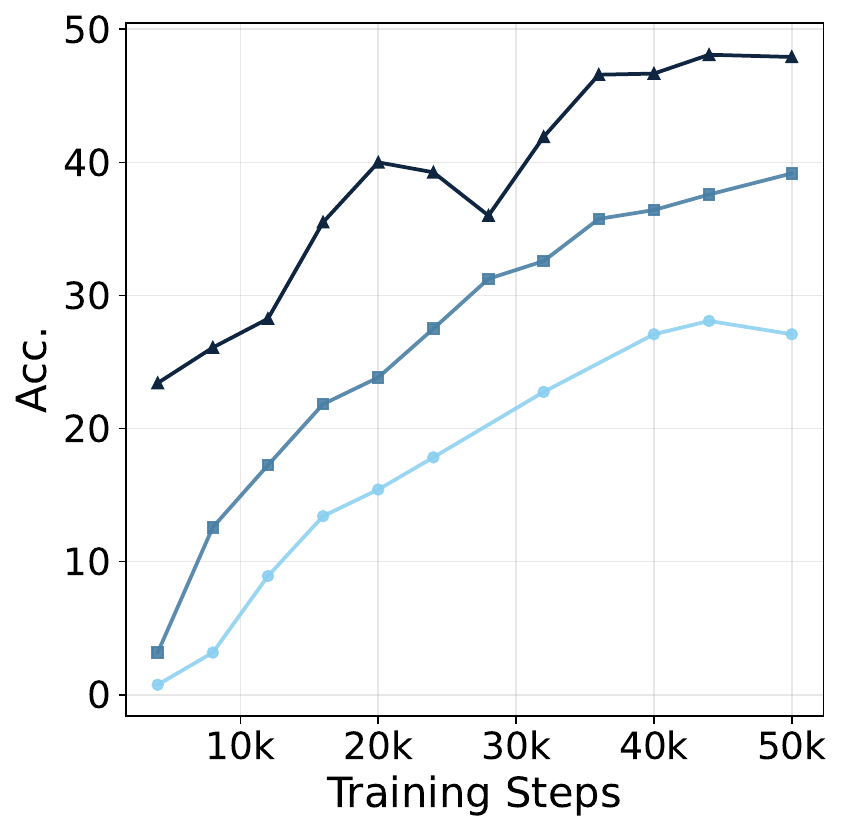}
         \caption{Competition Math \label{fig:7b_avg}}
     \end{subfigure}
     \hfill
    \begin{subfigure}[b]{0.24\textwidth}
         \centering
         \includegraphics[trim={0.3cm 0.25cm 0.2cm 0.2cm},clip,width=\textwidth]{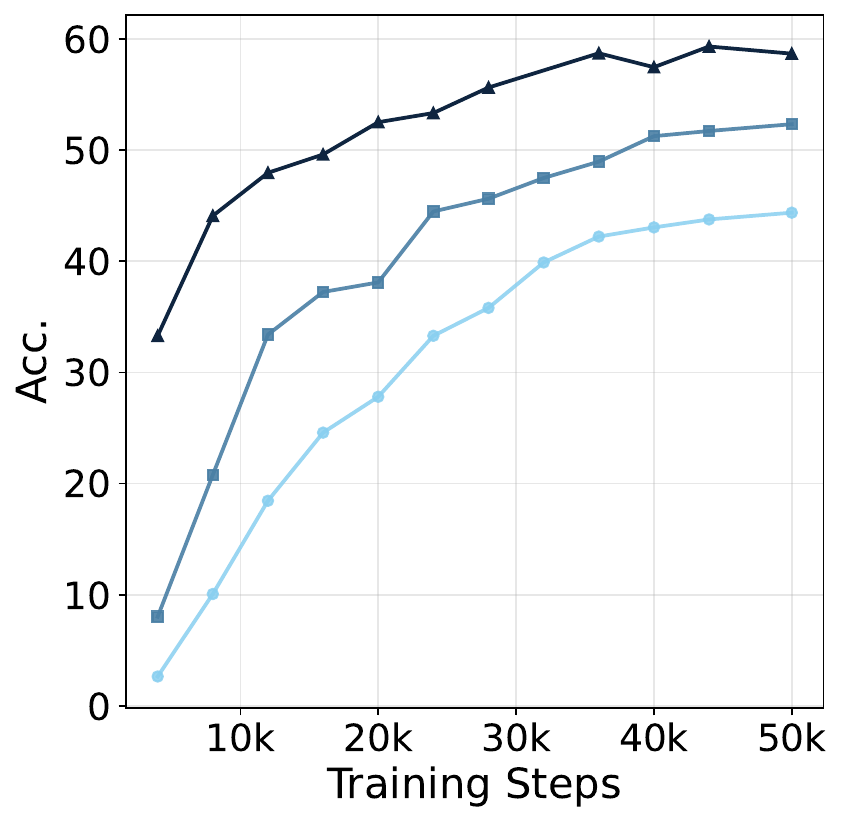}
         \caption{LCB V5 \label{fig:7b_lcb_v5}}
     \end{subfigure}
     \hfill
     \begin{subfigure}[b]{0.24\textwidth}
         \centering
         \includegraphics[trim={0.3cm 0.25cm 0.2cm 0.2cm},clip,width=\textwidth]{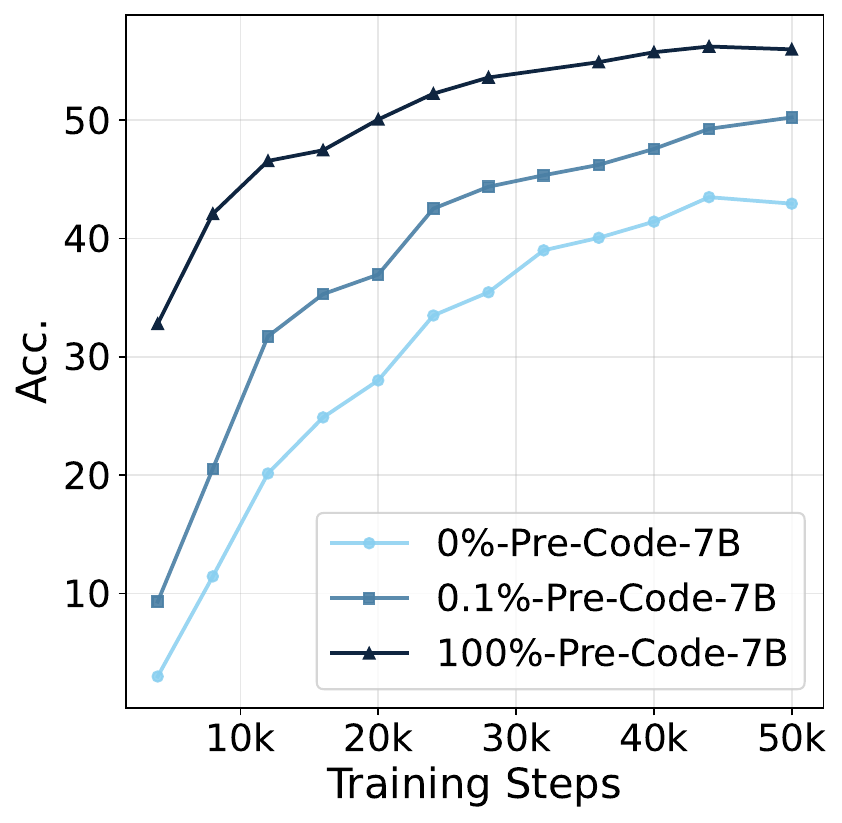}
         \caption{LCB V6 \label{fig:7b_lcb_v6}}
     \end{subfigure}
     \caption{SFT evaluation curves using different amounts of pre-training backbones (7B).\label{fig:7b}}
\end{figure*}

\begin{figure*}[h!]
     \centering
     \begin{subfigure}[b]{0.24\textwidth}
         \centering
         \includegraphics[trim={0.3cm 0.25cm 0.2cm 0.2cm},clip,width=\textwidth]{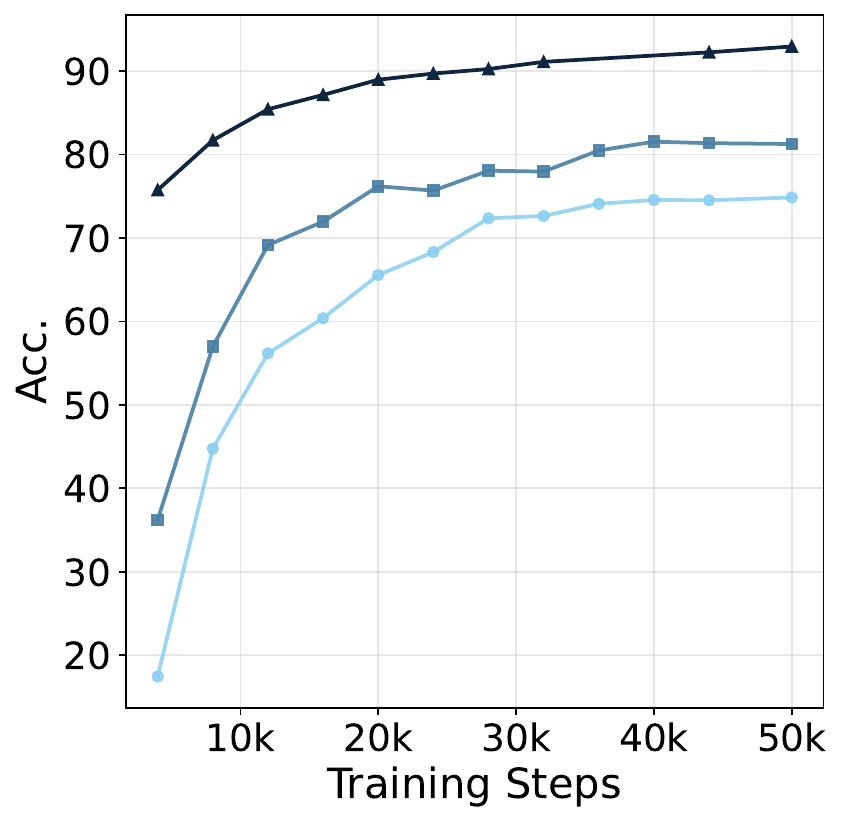}
         \caption{Math-500 \label{fig:3b_math}}
     \end{subfigure}
     \hfill
     \begin{subfigure}[b]{0.24\textwidth}
         \centering
         \includegraphics[trim={0.3cm 0.25cm 0.2cm 0.2cm},clip,width=\textwidth]{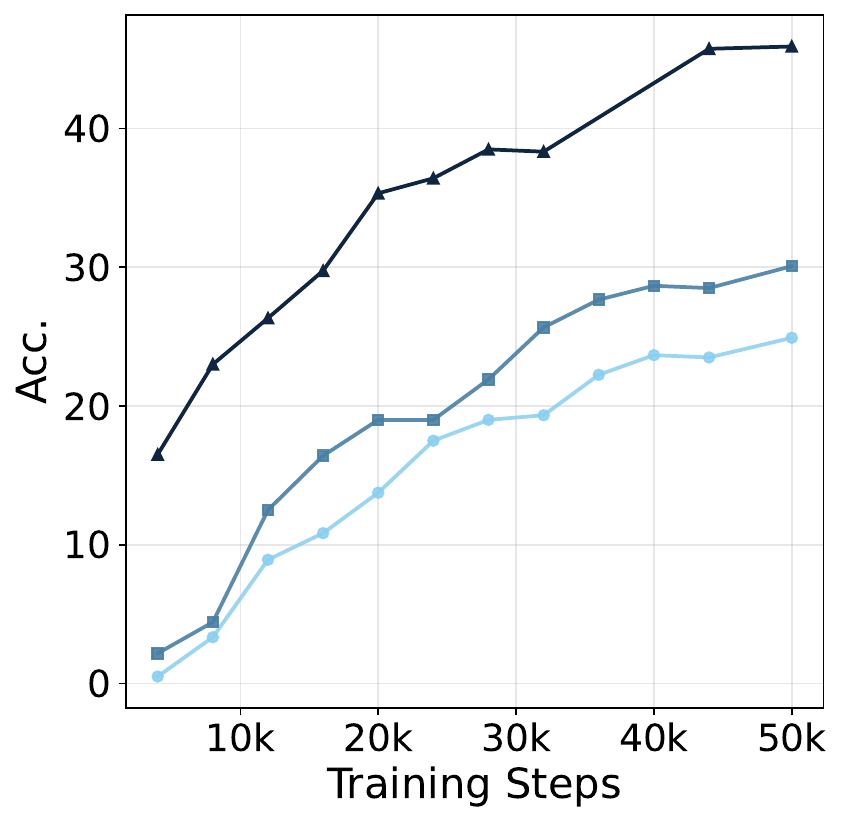}
         \caption{Competition Math \label{fig:3b_avg}}
     \end{subfigure}
     \hfill
    \begin{subfigure}[b]{0.24\textwidth}
         \centering
         \includegraphics[trim={0.3cm 0.25cm 0.2cm 0.2cm},clip,width=\textwidth]{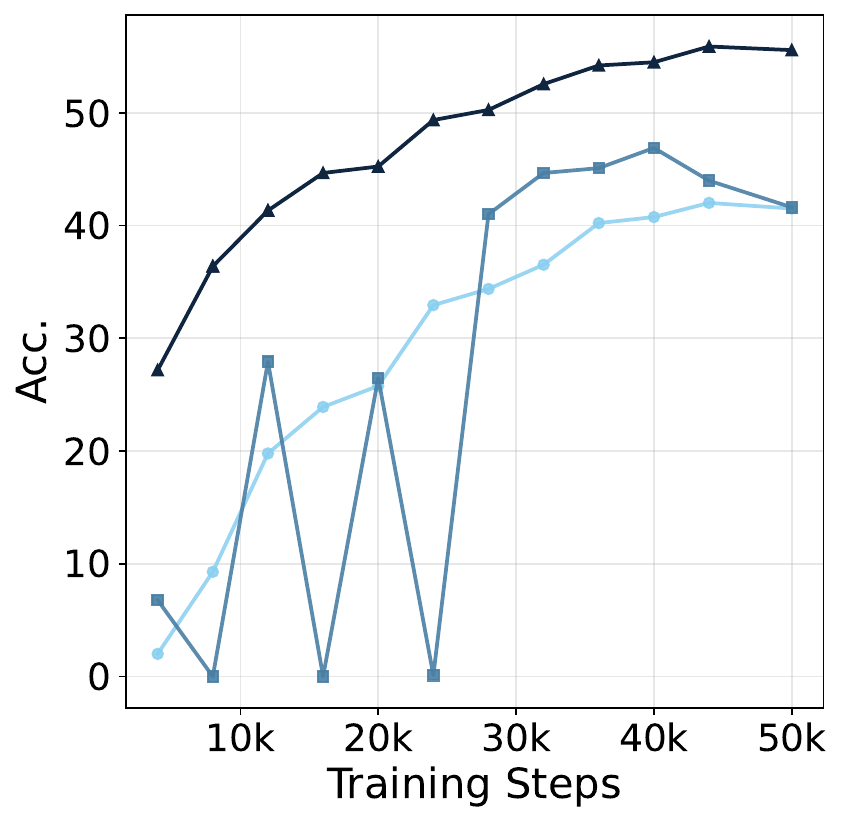}
         \caption{LCB V5 \label{fig:3b_lcb_v5}}
     \end{subfigure}
     \hfill
     \begin{subfigure}[b]{0.24\textwidth}
         \centering
         \includegraphics[trim={0.3cm 0.25cm 0.2cm 0.2cm},clip,width=\textwidth]{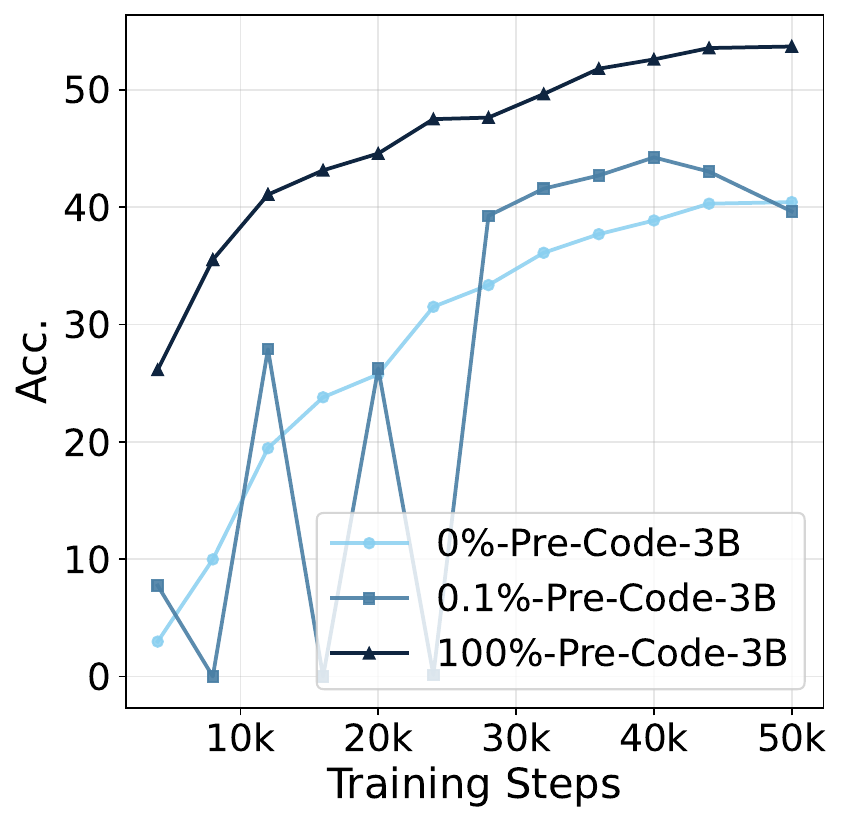}
         \caption{LCB V6 \label{fig:3b_lcb_v6}}
     \end{subfigure}
     \caption{SFT evaluation curves using different amounts of pre-training backbones (3B). \label{fig:3b}}
\end{figure*}

\begin{figure*}[h]
     \centering
     \begin{subfigure}[b]{0.24\textwidth}
         \centering
         \includegraphics[trim={0.3cm 0.25cm 0.2cm 0.2cm},clip,width=\textwidth]{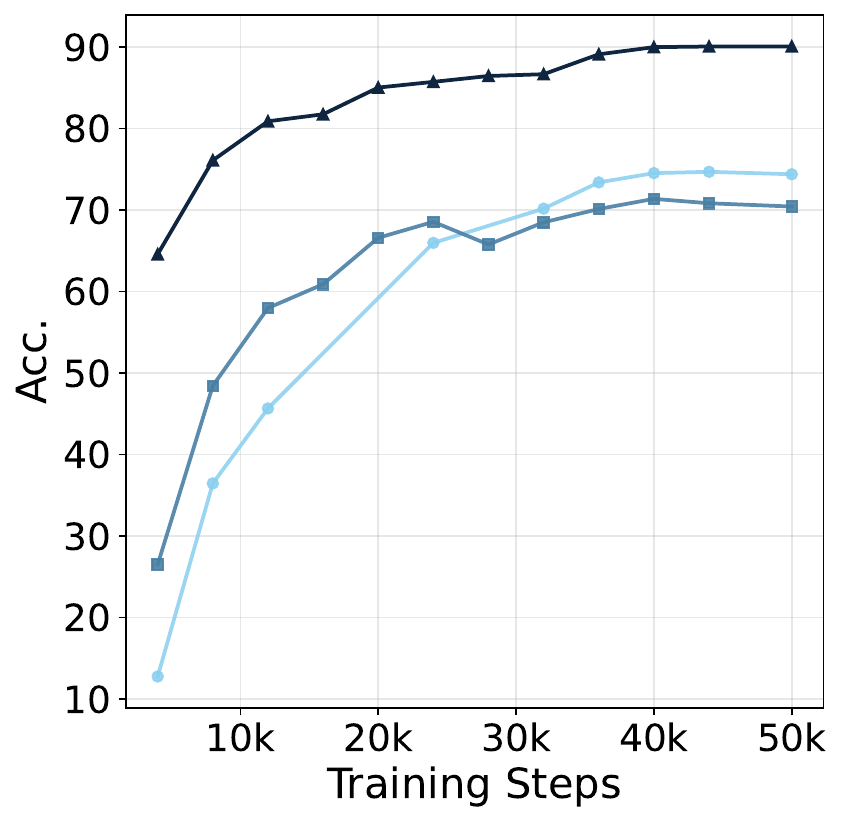}
         \caption{Math-500 \label{fig:1b_math}}
     \end{subfigure}
     \hfill
     \begin{subfigure}[b]{0.24\textwidth}
         \centering
         \includegraphics[trim={0.3cm 0.25cm 0.2cm 0.2cm},clip,width=\textwidth]{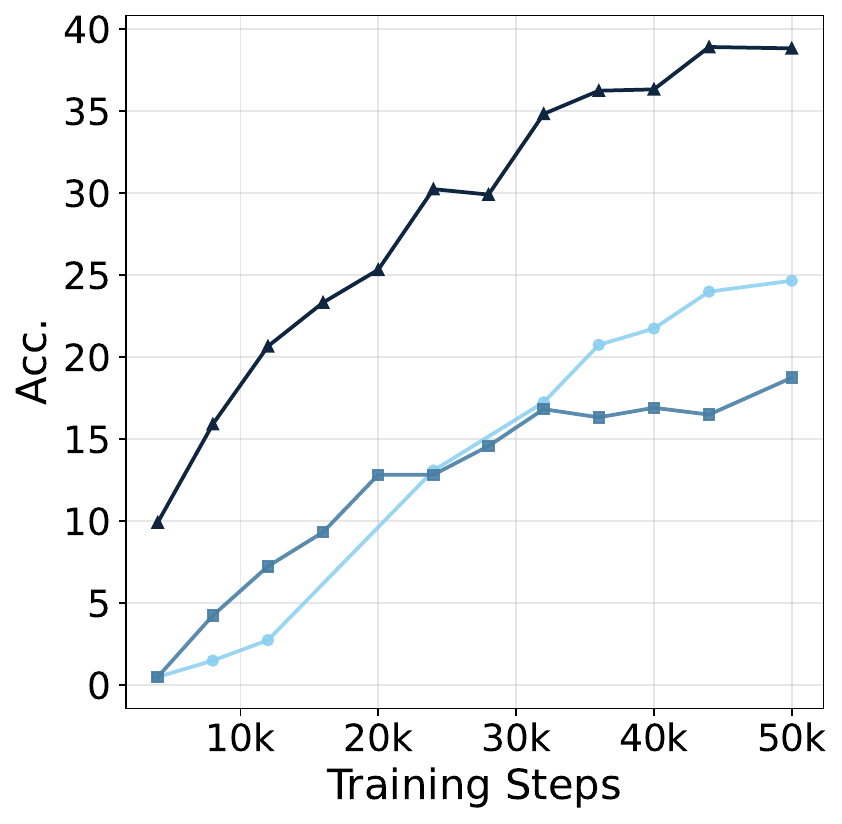}
         \caption{Competition Math \label{fig:1b_avg}}
     \end{subfigure}
     \hfill
    \begin{subfigure}[b]{0.24\textwidth}
         \centering
         \includegraphics[trim={0.3cm 0.25cm 0.2cm 0.2cm},clip,width=\textwidth]{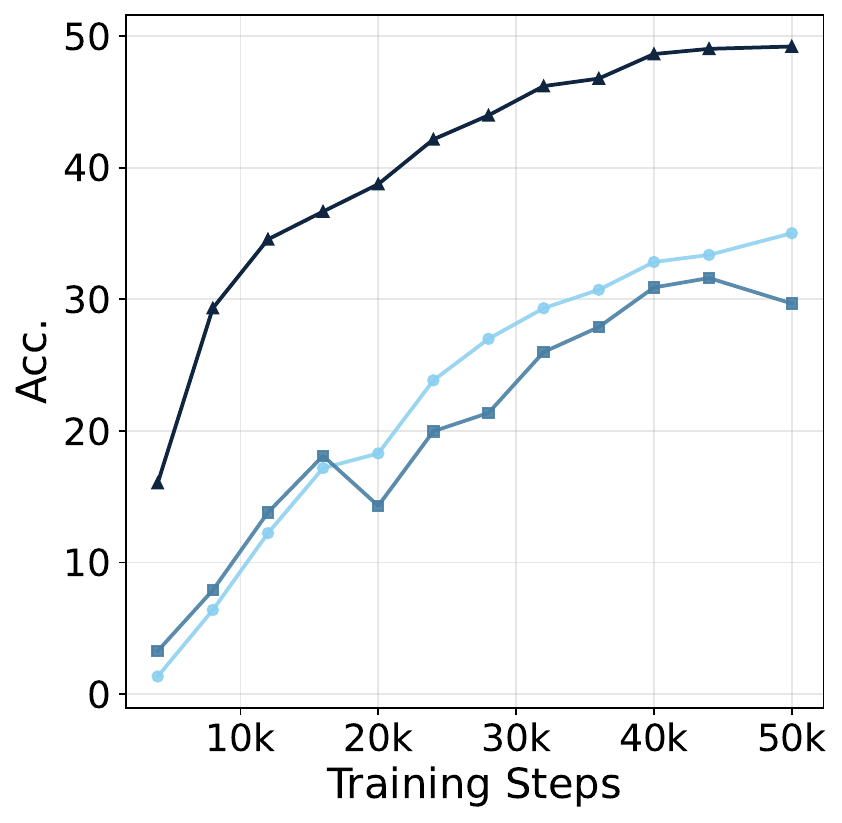}
         \caption{LCB V5 \label{fig:1b_lcb_v5}}
     \end{subfigure}
     \hfill
     \begin{subfigure}[b]{0.24\textwidth}
         \centering
         \includegraphics[trim={0.3cm 0.25cm 0.2cm 0.2cm},clip,width=\textwidth]{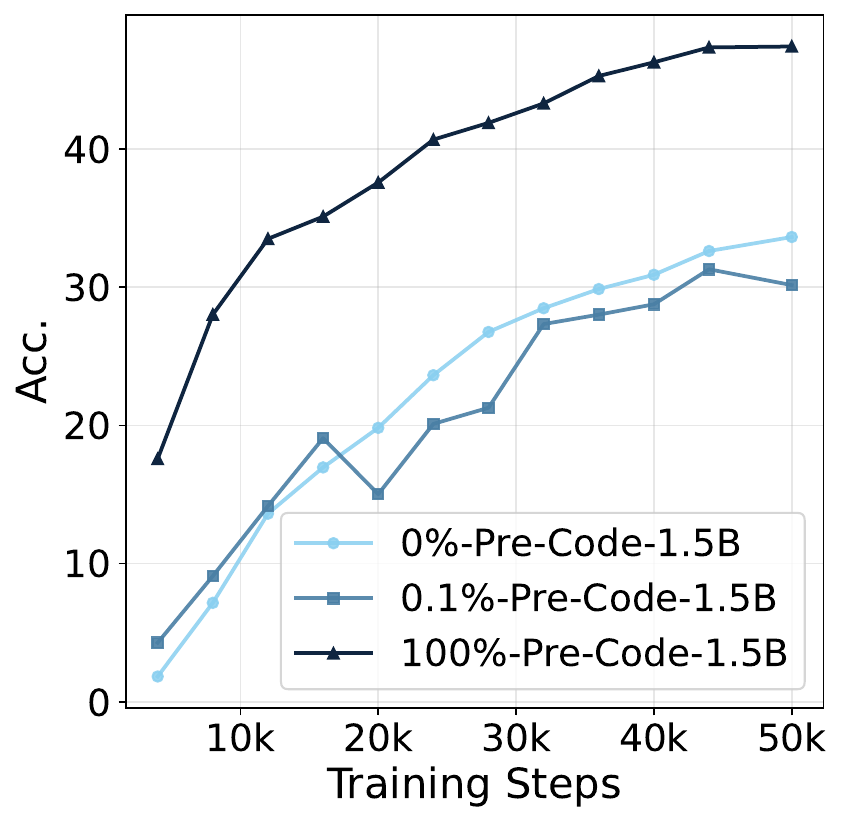}
         \caption{LCB V6 \label{fig:1b_lcb_v6}}
     \end{subfigure}
     \caption{SFT evaluation curves using different amounts of pre-training backbones (1.5B). \label{fig:1b}}
\end{figure*}

\cref{fig:7b}, \cref{fig:3b} and \cref{fig:1b} presents the performance curves of the 7B/3B/1.5B models, respectively, using different amounts of data for the pre-training backbones.

%% file: 95_math_perfomance_training.tex
\section{Performance During Training of Competition Math Benchmarks}
\label{app:math_curves}

\begin{figure*}[h]
     \centering
     \begin{subfigure}[b]{0.24\textwidth}
         \centering
         \includegraphics[trim={0.3cm 0.25cm 0.2cm 0.2cm},clip,width=\textwidth]{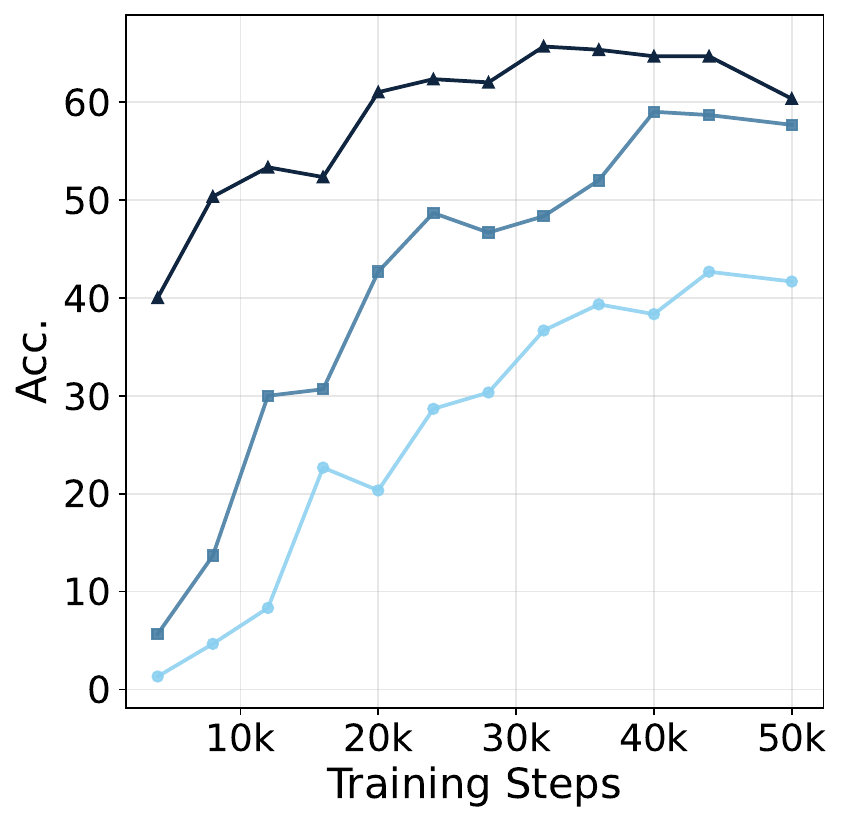}
         \caption{AIME 24 \label{fig:14b_aime
         _24}}
     \end{subfigure}
     \hfill
     \begin{subfigure}[b]{0.24\textwidth}
         \centering
         \includegraphics[trim={0.3cm 0.25cm 0.2cm 0.2cm},clip,width=\textwidth]{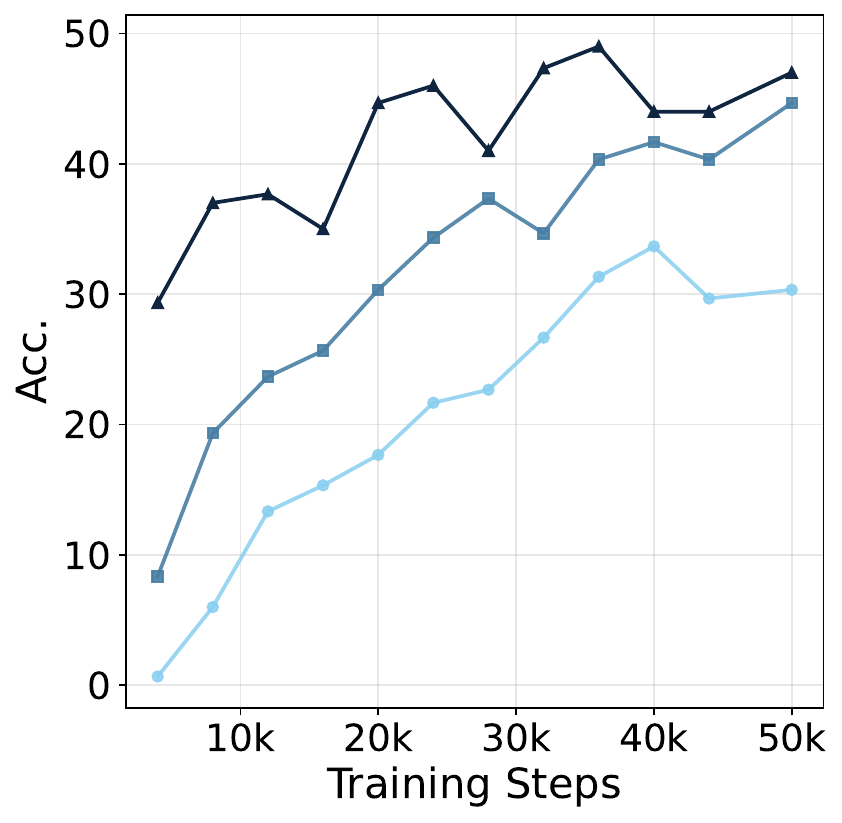}
         \caption{AIME 25 \label{fig:14b_aime
         _25}}
     \end{subfigure}
     \hfill
    \begin{subfigure}[b]{0.24\textwidth}
         \centering
         \includegraphics[trim={0.3cm 0.25cm 0.2cm 0.2cm},clip,width=\textwidth]{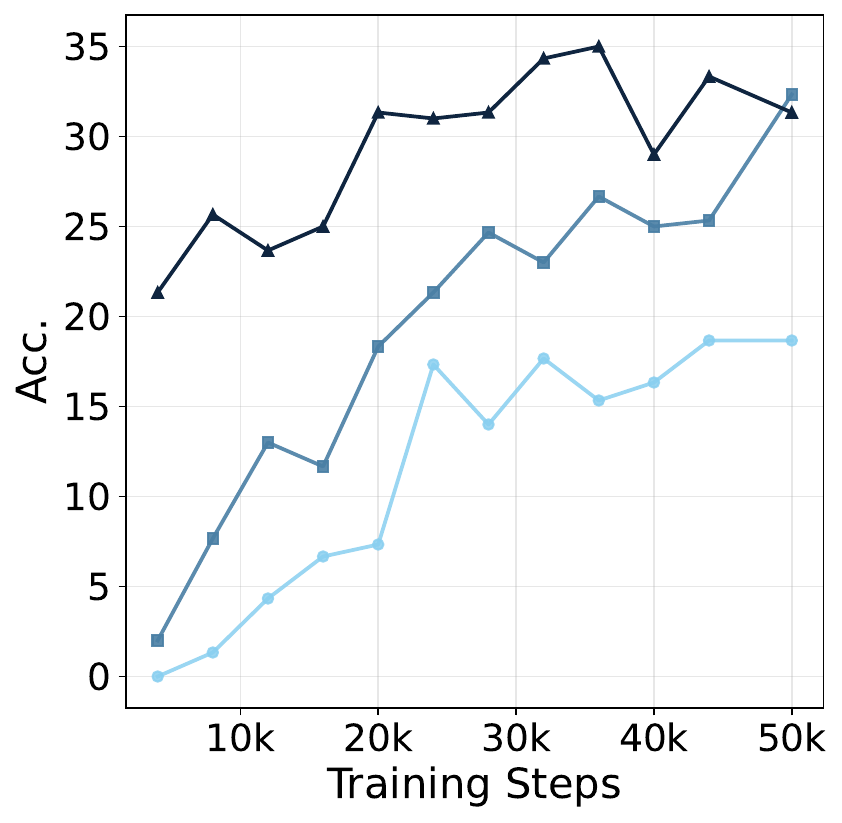}
         \caption{HMMT Feb 25 \label{fig:14b_hmmt
         _feb}}
     \end{subfigure}
     \hfill
     \begin{subfigure}[b]{0.24\textwidth}
         \centering
         \includegraphics[trim={0.3cm 0.25cm 0.2cm 0.2cm},clip,width=\textwidth]{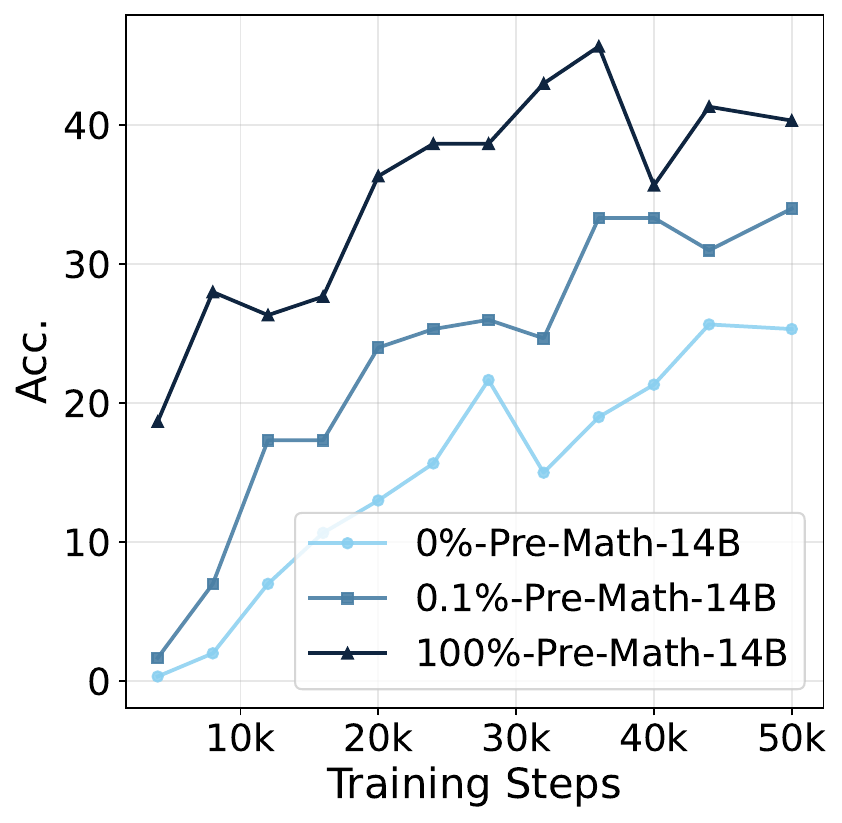}
         \caption{HMMT Nov 25 \label{fig:14b_hmmt
         _nov}}
     \end{subfigure}
     \caption{SFT evaluation curves over the math competition benchmarks, using different amounts of pre-training backbones (14B). \label{fig:14b_math_full}}
\end{figure*}

\begin{figure*}[h]
     \centering
     \begin{subfigure}[b]{0.24\textwidth}
         \centering
         \includegraphics[trim={0.3cm 0.25cm 0.2cm 0.2cm},clip,width=\textwidth]{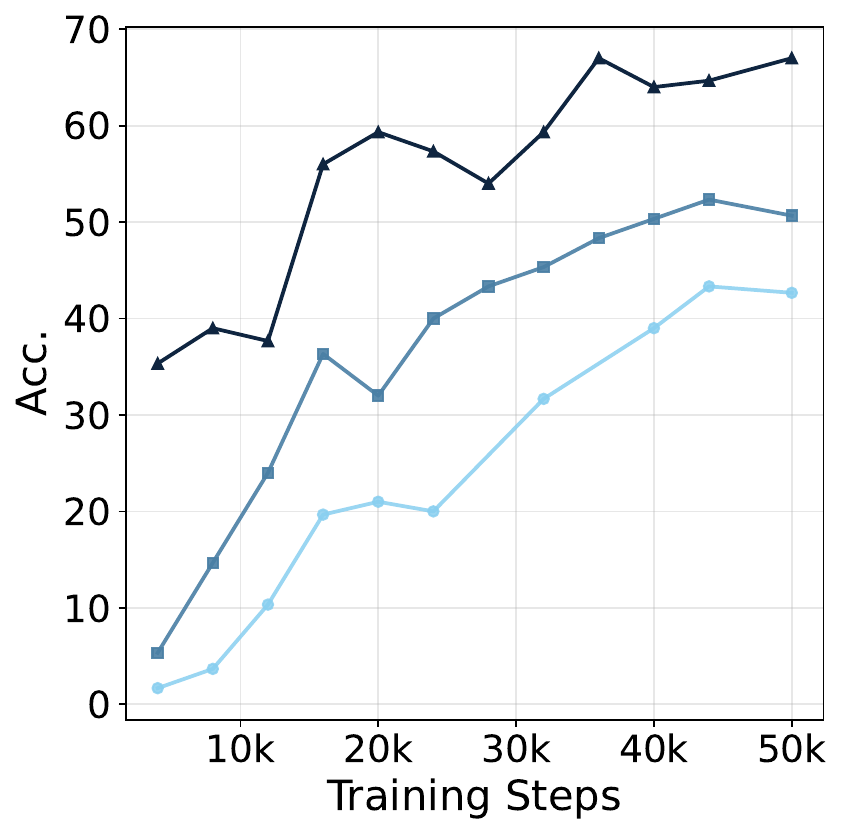}
         \caption{AIME 24 \label{fig:7b_aime
         _24}}
     \end{subfigure}
     \hfill
     \begin{subfigure}[b]{0.24\textwidth}
         \centering
         \includegraphics[trim={0.3cm 0.25cm 0.2cm 0.2cm},clip,width=\textwidth]{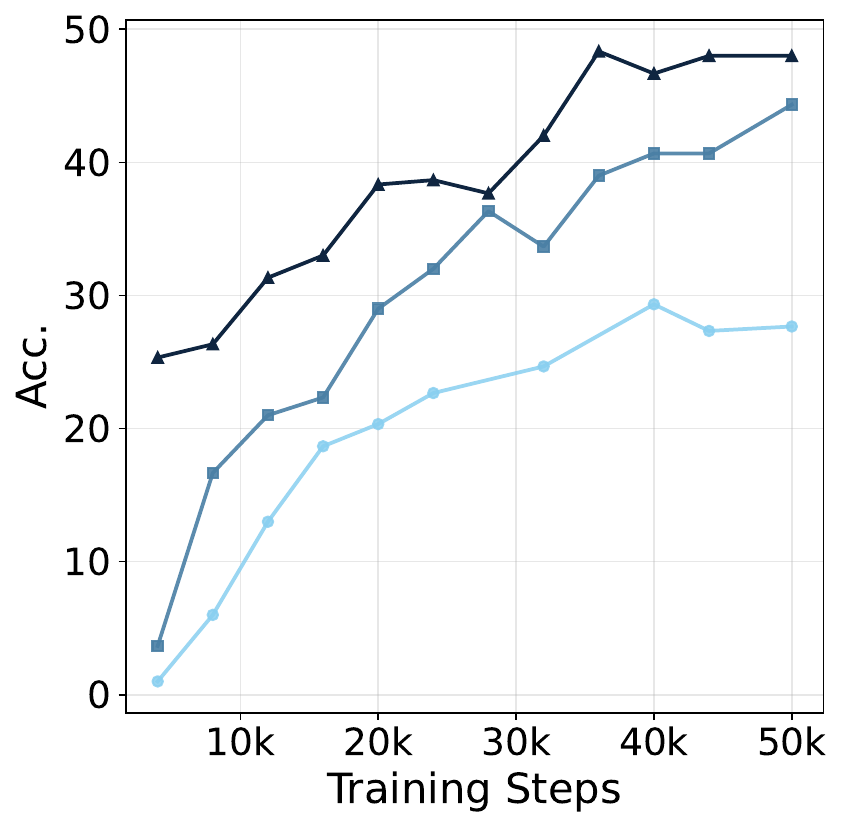}
         \caption{AIME 25 \label{fig:7b_aime
         _25}}
     \end{subfigure}
     \hfill
    \begin{subfigure}[b]{0.24\textwidth}
         \centering
         \includegraphics[trim={0.3cm 0.25cm 0.2cm 0.2cm},clip,width=\textwidth]{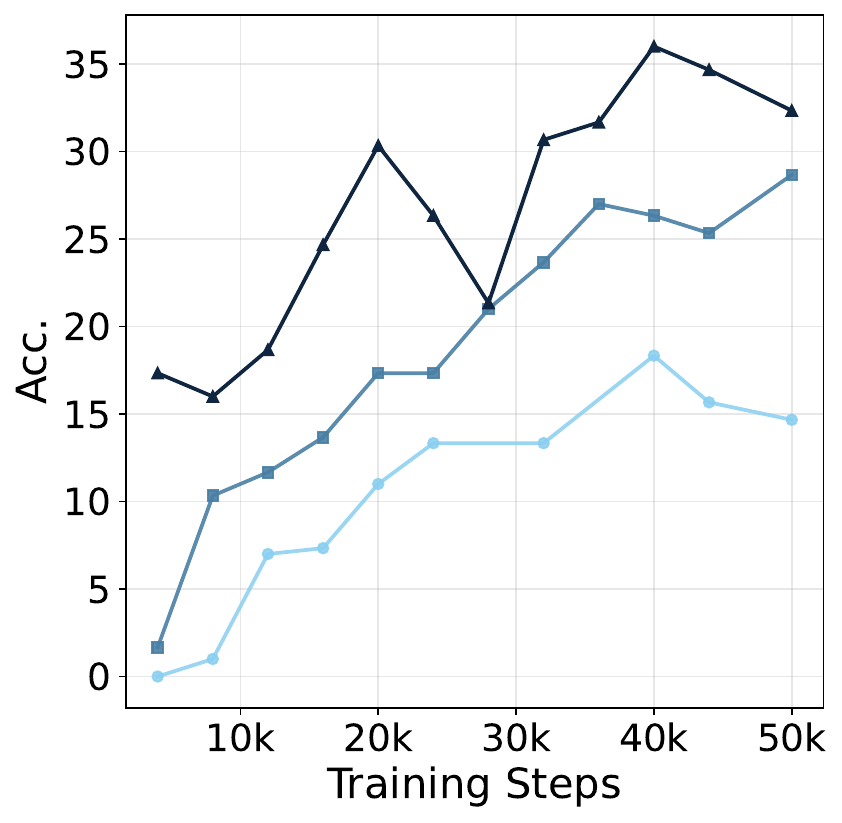}
         \caption{HMMT Feb 25 \label{fig:7b_hmmt
         _feb}}
     \end{subfigure}
     \hfill
     \begin{subfigure}[b]{0.24\textwidth}
         \centering
         \includegraphics[trim={0.3cm 0.25cm 0.2cm 0.2cm},clip,width=\textwidth]{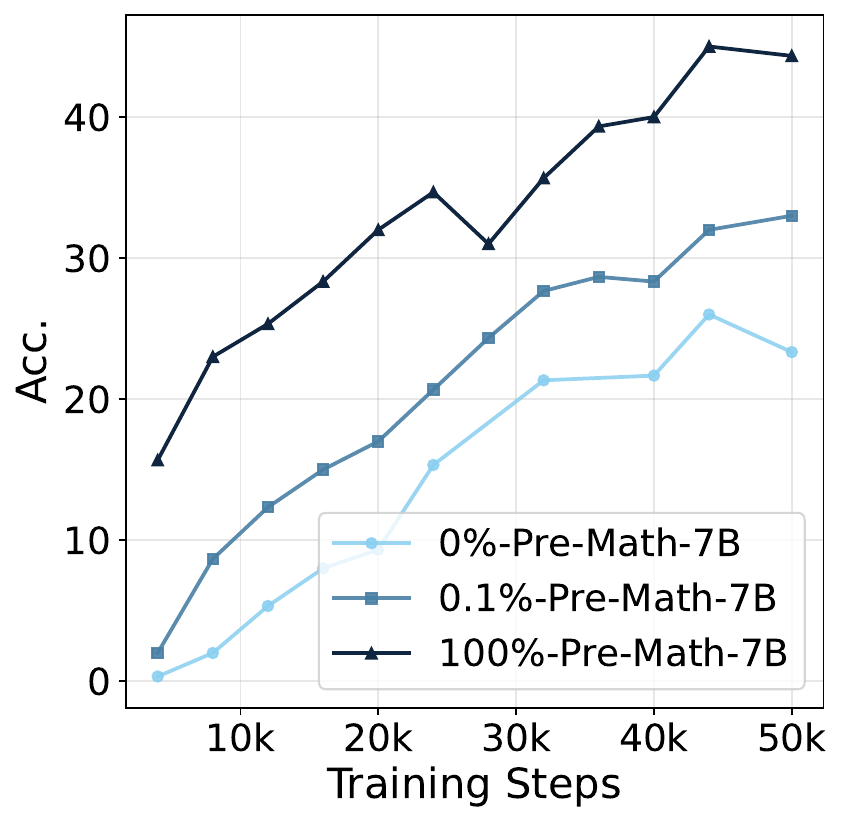}
         \caption{HMMT Nov 25 \label{fig:7b_hmmt
         _nov}}
     \end{subfigure}
     \caption{SFT evaluation curves over the math competition benchmarks, using different amounts of pre-training backbones (7B). \label{fig:7b_math_full}}
\end{figure*}

\begin{figure*}[h]
     \centering
     \begin{subfigure}[b]{0.24\textwidth}
         \centering
         \includegraphics[trim={0.3cm 0.25cm 0.2cm 0.2cm},clip,width=\textwidth]{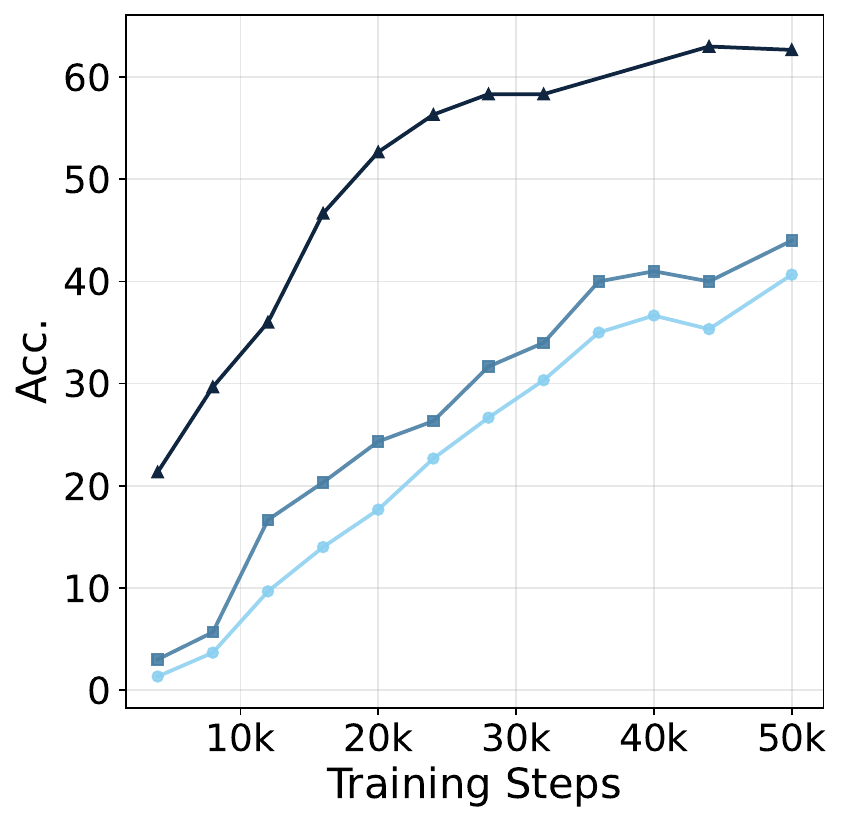}
         \caption{AIME 24 \label{fig:3b_aime
         _24}}
     \end{subfigure}
     \hfill
     \begin{subfigure}[b]{0.24\textwidth}
         \centering
         \includegraphics[trim={0.3cm 0.25cm 0.2cm 0.2cm},clip,width=\textwidth]{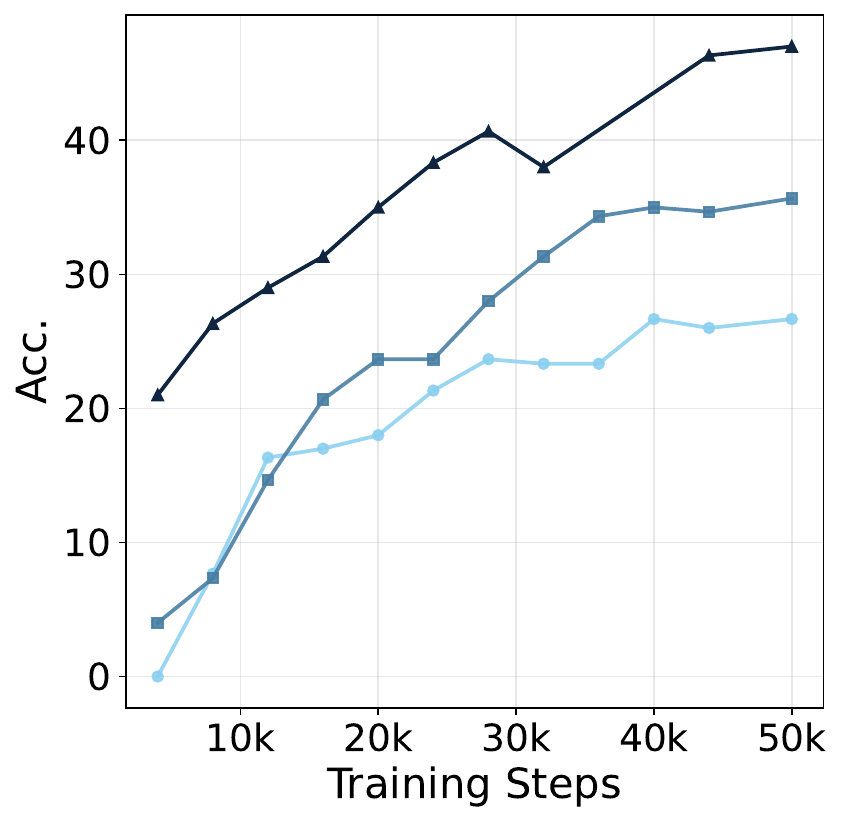}
         \caption{AIME 25 \label{fig:3b_aime
         _25}}
     \end{subfigure}
     \hfill
    \begin{subfigure}[b]{0.24\textwidth}
         \centering
         \includegraphics[trim={0.3cm 0.25cm 0.2cm 0.2cm},clip,width=\textwidth]{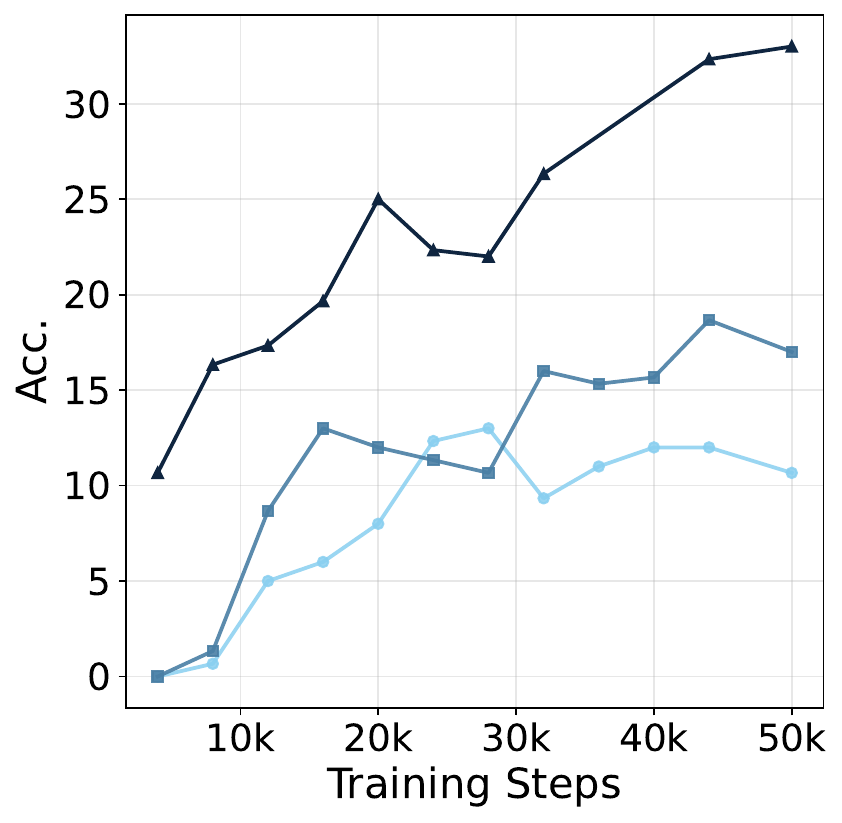}
         \caption{HMMT Feb 25 \label{fig:3b_hmmt
         _feb}}
     \end{subfigure}
     \hfill
     \begin{subfigure}[b]{0.24\textwidth}
         \centering
         \includegraphics[trim={0.3cm 0.25cm 0.2cm 0.2cm},clip,width=\textwidth]{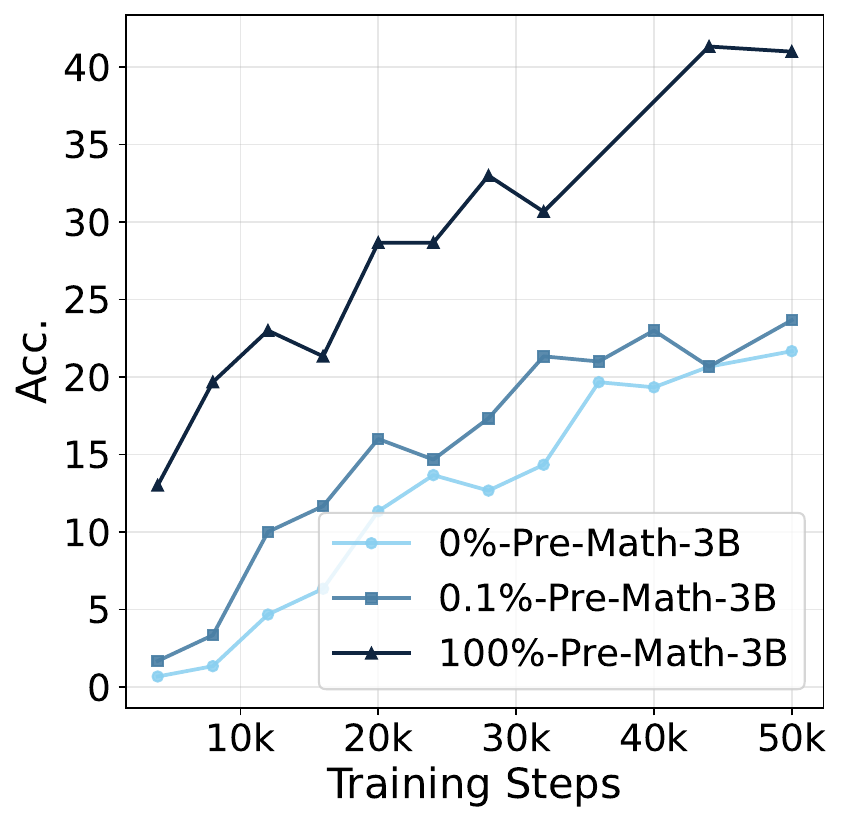}
         \caption{HMMT Nov 25 \label{fig:3b_hmmt
         _nov}}
     \end{subfigure}
     \caption{SFT evaluation curves over the math competition benchmarks, using different amounts of pre-training backbones (3B). \label{fig:3b_math_full}}
\end{figure*}

\begin{figure*}[h]
     \centering
     \begin{subfigure}[b]{0.24\textwidth}
         \centering
         \includegraphics[trim={0.3cm 0.25cm 0.2cm 0.2cm},clip,width=\textwidth]{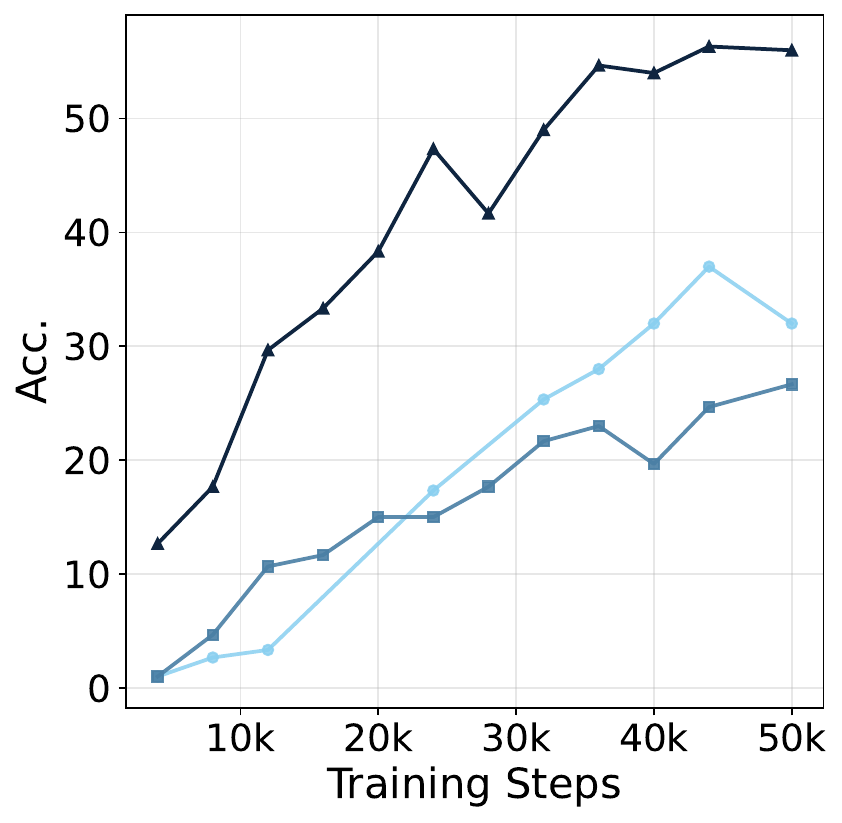}
         \caption{AIME 24 \label{fig:1b_aime
         _24}}
     \end{subfigure}
     \hfill
     \begin{subfigure}[b]{0.24\textwidth}
         \centering
         \includegraphics[trim={0.3cm 0.25cm 0.2cm 0.2cm},clip,width=\textwidth]{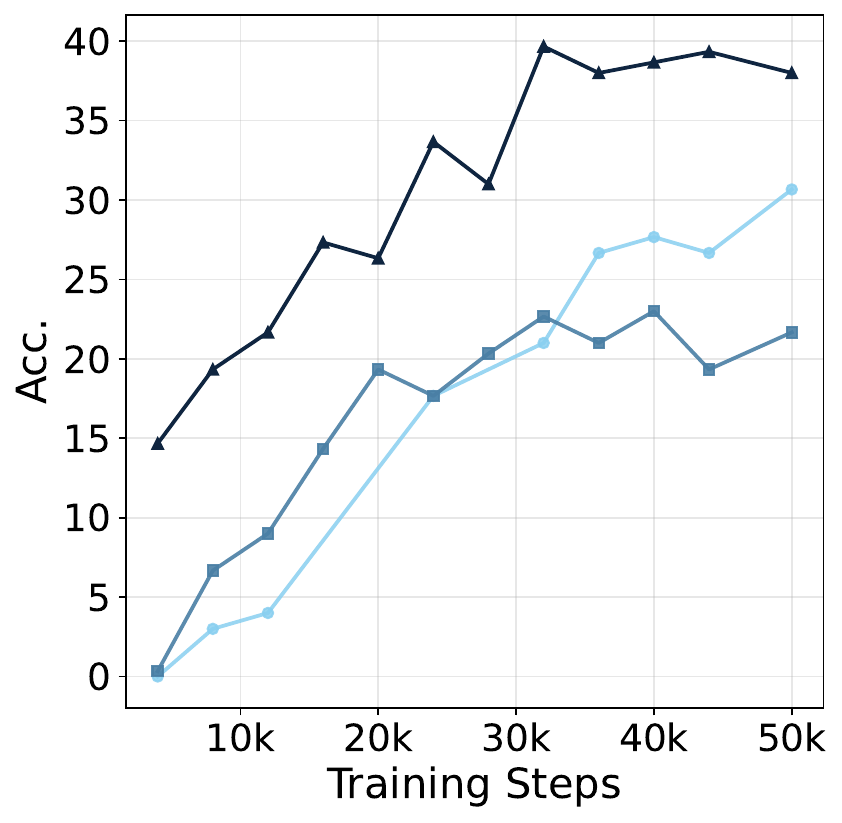}
         \caption{AIME 25 \label{fig:1b_aime
         _25}}
     \end{subfigure}
     \hfill
    \begin{subfigure}[b]{0.24\textwidth}
         \centering
         \includegraphics[trim={0.3cm 0.25cm 0.2cm 0.2cm},clip,width=\textwidth]{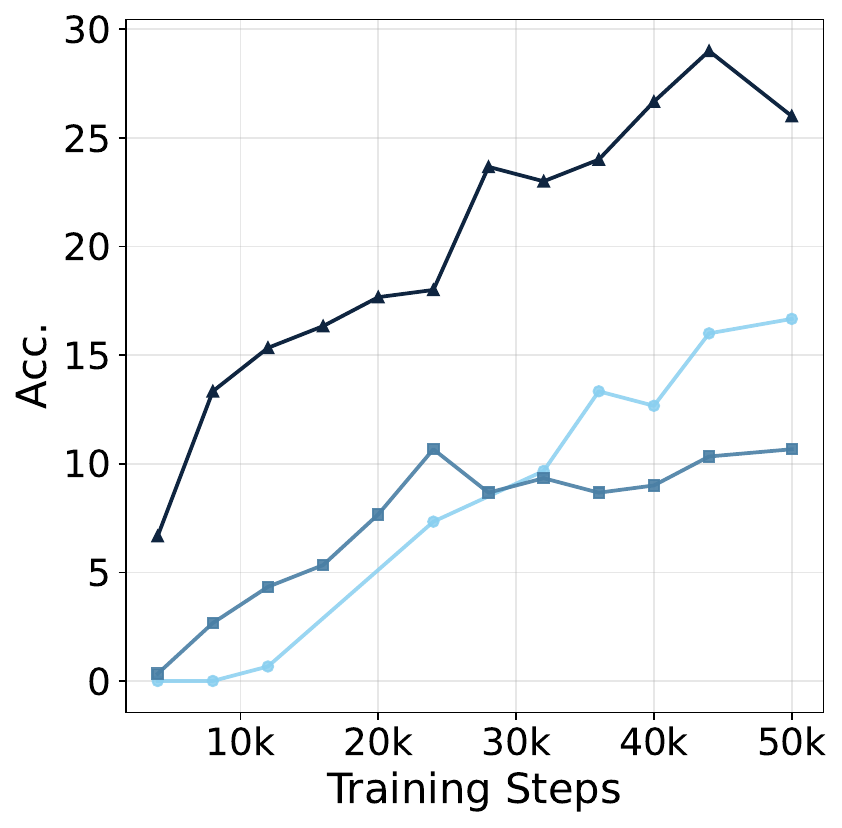}
         \caption{HMMT Feb 25 \label{fig:1b_hmmt
         _feb}}
     \end{subfigure}
     \hfill
     \begin{subfigure}[b]{0.24\textwidth}
         \centering
         \includegraphics[trim={0.3cm 0.25cm 0.2cm 0.2cm},clip,width=\textwidth]{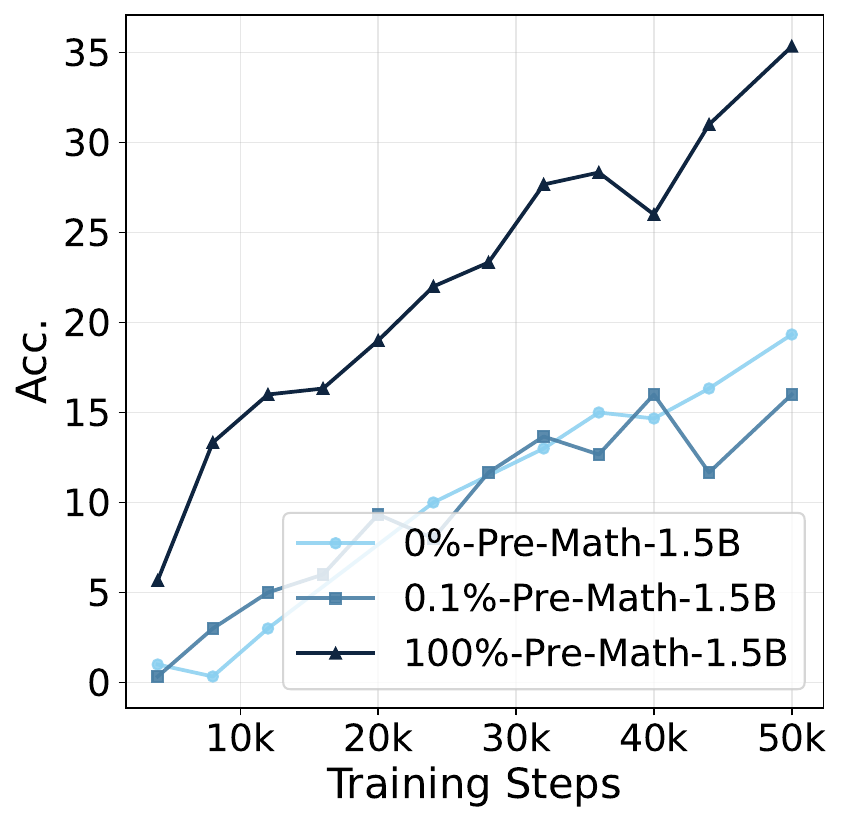}
         \caption{HMMT Nov 25 \label{fig:1b_hmmt
         _nov}}
     \end{subfigure}
     \caption{SFT evaluation curves over the math competition benchmarks, using different amounts of pre-training backbones (1.5B). \label{fig:1b_math_full}}
\end{figure*}

We present the performance curves of the 14/7/3/1.5B models for the competition math benchmarks in \cref{fig:14b_math_full}, \cref{fig:7b_math_full}, \cref{fig:3b_math_full} and \cref{fig:1b_math_full}, respectively.

